%% file: ijcai26.tex
\documentclass{article}
\usepackage{ijcai26}

\usepackage{times}
\usepackage{soul}
\usepackage{url}
\usepackage[hidelinks]{hyperref}
\usepackage[utf8]{inputenc}
\usepackage[small]{caption}
\usepackage{graphicx}
\usepackage{amsmath}
\usepackage{amsthm}
\usepackage{booktabs}
\usepackage{algorithm}
\usepackage{algorithmic}
\usepackage[switch]{lineno}

\usepackage{xurl}
\usepackage{url}            
\usepackage{booktabs}       
\usepackage{amsfonts}       
\usepackage{nicefrac}       
\usepackage{microtype}      
\usepackage{enumitem}
\usepackage{graphicx}
\usepackage{tcolorbox}
\usepackage{multirow}
\usepackage{makecell}
\usepackage{color}
\usepackage{enumitem}
\usepackage{amsmath}
\usepackage[table]{xcolor}
\usepackage{bm}
\usepackage{pifont}

\usepackage{trajan}
\usepackage{CJKutf8}
\usepackage{wrapfig}

\usepackage{tcolorbox}
\tcbuselibrary{breakable}
\usepackage{xspace}

\usepackage{pifont}

\usepackage{float}

\definecolor{mygray}{HTML}{333333}

\renewcommand{\tiny}{\fontsize{7.5pt}{7.5pt}\selectfont}
\renewcommand{\footnotesize}{\fontsize{8pt}{8pt}\selectfont}
\renewcommand{\small}{\fontsize{8.5pt}{8.5pt}\selectfont}

\newcommand{\sysone}{{\textit{system 1}}\xspace}
\newcommand{\systwo}{{\textit{system 2}}\xspace}
\newcommand{\bench}{{S1-Bench}\xspace}

\newcommand{\cmark}{{\color{kellygreen} \ding{51}}}
\newcommand{\xmark}{{\color{alizarin} \ding{55}}}

\definecolor{kellygreen}{rgb}{0.3, 0.73, 0.09}
\definecolor{alizarin}{rgb}{0.82, 0.1, 0.26}
\definecolor{burgundy}{RGB}{168, 0, 50}
\definecolor{whitegray}{RGB}{200, 200, 200}

\hypersetup{
    colorlinks=True,       
    pdfborder={0 0 0},      
    bookmarksnumbered=true  
}

\title{Exploring the System 1 Thinking Capability of Large Reasoning Models}

\author{
Wenyuan Zhang$^{1,2}$\thanks{~denotes equal contribution. $^\dagger$ denotes corresponding author.}
\and
Shuaiyi Nie$^{1,2*}$\and
Xinghua Zhang$^3$\and
Zefeng Zhang$^{1,2}$\and
Tingwen Liu$^{1,2\dagger}$\\
\affiliations
$^1$Institute of Information Engineering, Chinese Academy of Sciences\\
$^2$School of Cyber Security, University of Chinese Academy of Sciences\\
$^3$Tongyi Lab, Alibaba Group\\
\emails
\{zhangwenyuan, nieshuaiyi, zhangzefeng, liutingwen\}@iie.ac.cn,\\
zhangxinghua.zxh@alibaba-inc.com
}

\begin{document}

\maketitle

\input{sections/abstract}

\input{sections/introduction}

\input{sections/method}

\input{sections/experiment}
\input{sections/exp_gut}

\input{sections/related-work}
\input{sections/conclusion}

\input{sections/acknowledgments}

\bibliographystyle{named}
\bibliography{ijcai26}

\clearpage
\appendix

\input{sections/appendix}

\end{document}

%% file: sections/abstract.tex
\begin{abstract}

This paper explores the \sysone thinking capability of Large Reasoning Models (LRMs), the intuitive ability to respond efficiently with minimal token usage.
While existing LRMs rely on long-chain reasoning and excel at complex tasks, their \sysone thinking ability remains largely underexplored.
This capability is essential as it reflects models' difficulty awareness and reasoning efficiency, both critical for real-world applications.
We propose S1-Bench, a multi-domain, multilingual benchmark comprising model-simple \sysone questions.
Our investigation of 28 LRMs reveals under-accuracy and inefficiency on \sysone problems.
We find existing efficient reasoning methods either generalize poorly to simple questions or sacrifice performance for efficiency.
Further exploration uncovers LRMs' early difficulty awareness accompanied by lower confidence, and shows that problem difficulty is implicitly encoded in hidden states\footnote{The code can be found in \url{https://github.com/WYRipple/S1_Bench}.}.

\end{abstract}

%
%
%

%% file: sections/introduction.tex
\section{Introduction}

Recent advances in Large Reasoning Models (LRMs), notably OpenAI’s o1/o3~\cite{openai-o1} and the DeepSeek-R1~\cite{guo2025deepseek} series, have propelled the development of Large Language Models (LLMs).
%
%
Unlike traditional LLMs that exhibit intuitive, heuristic \sysone thinking, LRMs demonstrate deliberate and analytical \systwo reasoning~\cite{qu2025survey,li2025survey} by explicitly generating external chain-of-thought~\cite{cot} before producing final answers.
%
%

%

While most prior research focuses on enhancing LRMs’ reasoning capabilities to achieve superior performance on complex tasks requiring \systwo ~\cite{yang2025qwen3,strategies-li,strategies-yeo}, exploration of their \sysone thinking abilities remains limited.
Specifically, \sysone thinking refers to the human-like intuitive ability~\cite{kahneman2011thinking} to answer questions rapidly with minimal tokens usage. 
This capability reflects a model’s perception of task difficulty as well as its proficiency in efficiently responding to questions it recognizes as simple~\cite{booch2021thinking}.
In real-world scenarios, where the vast majority of user queries are relatively straightforward for models, ensuring accurate and efficient \sysone thinking is particularly crucial~\cite{yang2024harnessing}.


In this work, we aim to explore the \sysone thinking capabilities of LRMs, yet a suitable benchmark remains absent.
Some studies attempt to evaluate LRMs using questions that humans consider simple (e.g., GSM8K-zero and RoR~\cite{chiang-lee-2024-reasoning,yan2025recitation}), yet these questions remain challenging for models.
%
%
Other works utilize single-domain benchmarks, such as basic mathematics (GSM8K and ASDIV~\cite{GSM8K,ASDIV}), but lack domain diversity.
%
Other research on LRM efficiency primarily targets challenging problems, such as AIME.
As shown in Table~\ref{tab:motivation}, the top six benchmarks which are most commonly used in these work remain difficult for small LLMs, making it unsuitable for exploring \sysone thinking capabilities.
%


\input{tables/motivation}


\input{figure/st_bench_data_construction}

To complement the exploration foundation, we introduce the \underline{S}ystem \underline{1} Thinking Capability \underline{Bench}mark (\textbf{S1-Bench}), which measures the performance of LRMs across various simple tasks that commonly encountered in real-world applications.
%
S1-Bench has the following three characteristics:
%
\textbf{(1) \textit{Simple}}. The questions can be easily answered by LLMs. As shown in Table~\ref{tab:motivation}, LLMs with 7-9B parameters can robustly provide correct answers across different temperatures.
%
\textbf{(2) \textit{Diverse}}. It encompasses four major categories and 28 subcategories in two languages.
\textbf{(3) \textit{Natural}}. The questions are clear, without any misleading elements or ambiguities, ensuring they can be answered intuitively.


We conduct extensive exploration on S1-Bench across 28 LRMs, yielding the following key findings:
(1) \textbf{Under-accuracy}. Despite employing deep reasoning, several LRMs exhibit under-accuracy on simple questions, particularly in knowledge-based and instruction-following tasks, suggesting that long-chain reasoning may introduce catastrophic forgetting.
(2) \textbf{Inefficiency}: Current LRMs lack \sysone capabilities across all categories of questions, with average output lengths 13.8$\times$ longer than small LLMs on \bench.
(3) \textbf{Limitations of efficient reasoning methods}: Existing efficient reasoning methods either suffer from limited generalization, failing to effectively compress length on \bench, or fail to balance performance and efficiency, leading to significant drops in accuracy.
(4) \textbf{Early awareness of problem difficulty}:
%
We observe that LRMs often exhibit explicit difficulty-related comments prior to formal reasoning, indicating that models possess an early awareness of problem difficulty.
Moreover, even when identifying a question as simple, LRMs still generate redundant responses with higher average token entropy, indicating stronger exploratory.
These reveal a gap between models’ difficulty awareness and reasoning behavior.
Furthermore, our representation-level analysis shows that question difficulty is implicitly encoded in model hidden states, as a simple single-layer MLP can capture difficulty-related signals.

Our contributions can be summarized as follows:
\begin{itemize}[nosep, topsep=2pt]
    \item We propose S1-Bench, a benchmark dedicated to evaluating LRMs' \sysone thinking capabilities, and a semi-automated construction workflow for \sysone questions.
    \item We reveal LRMs' under-accuracy and inefficiency in \sysone thinking, and highlight the inadequacy of current efficient reasoning methods, informing future research.
    \item We explore LRMs' early difficulty awareness, revealing correlations between explicit responses and model confidence, and disentanglement of difficulty in hidden states.

\end{itemize}

%

%

%% file: tables/motivation.tex
\begin{table}[t!]
\centering
\footnotesize
\renewcommand{\arraystretch}{1.0}
\setlength\tabcolsep{4.5pt}
    \begin{tabular}{l | c c c | c }
    \toprule
    \multirow{2}{*}{\raisebox{-0.1\height}{\textbf{Benchmark}}} & Cross & Realistic & Multi- & \multirow{2}{*}{\raisebox{-0.1\height}{Acc.}} \\
     & Domain & Scenarios & lingual &  \\
    \midrule
    AIME & \xmark & \cmark & \xmark & 6.67  \\
    GPQA & \cmark & \cmark & \xmark & 24.94  \\
    Olympiad & \cmark & \cmark & \cmark & 27.94  \\ 
    AMC & \xmark & \cmark & \xmark & 31.88  \\
    MATH500 & \xmark & \cmark & \xmark & 58.30  \\ 
    MMLU  & \cmark & \cmark & \xmark & 66.27  \\ 
    \midrule
    GSM8K & \xmark & \cmark & \xmark & 87.45  \\ 
    ASDIV  & \xmark & \cmark & \xmark & 97.51  \\ 
    \midrule
    GSM8K-zero  & \xmark & \xmark & \xmark & 77.98  \\ 
    RoR-Bench  & \xmark & \xmark & \xmark & 14.24  \\ 
    \midrule
    \textbf{S1-Bench (ours)} & \cmark  & \cmark & \cmark & \textbf{100.00} \\ 
\bottomrule
\end{tabular}
\caption{Characteristics of S1-Bench, with `Acc.' representing the average accuracy of four 7-9B LLMs. See {Appendix A.1} for more details.}
\label{tab:motivation}
\end{table}

%% file: figure/st_bench_data_construction.tex
\begin{figure*}[!t]  
\centering  
\includegraphics[width=16cm]{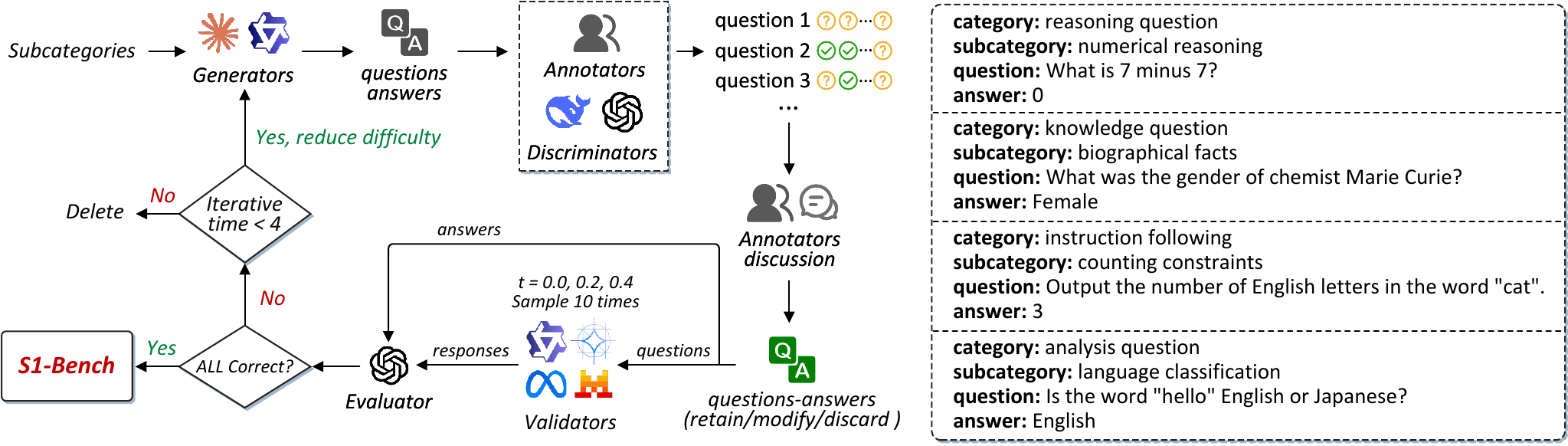}
\caption{Construction workflow for \bench and an illustrative example from each major category.}
\label{fig:get_data}
\end{figure*}

%% file: sections/method.tex
\section{\bench}

We introduce \bench, a bilingual, multi-domain benchmark designed to evaluate \sysone thinking capability of LRM on extremely simple questions.
%
These questions are easily solvable by traditional LLMs.
\bench, which covers both \textit{English} and \textit{Chinese}, is organized into four major categories: \textit{reasoning} (\textsc{Rsn}), \textit{knowledge} (\textsc{Kno}), \textit{instruction following} (\textsc{If}) and \textit{analysis} (\textsc{Ana}), representing major dimensions commonly employed in LLM capability evaluation~\cite{zheng2023judging,chang2024survey}.
This section begins with how simplicity is ensured, then the detailed construction workflow for \bench, and concludes with an overview of the dataset statistics.
Figure~\ref{fig:get_data} shows the construction workflow and an illustrative example per category.

\subsection{How to Ensure Simplicity?}
We ensure questions are simple and suitable for \sysone thinking through the following two aspects.
\subsubsection{A Priori Simplicity Constraints}
\label{sec:priori-simple}

We begin by generating question–answer pairs through collaboration between humans and LLMs.
Each pair is required to satisfy both the general and the category-specific simplicity criteria.
The general simplicity criteria requires that:
(1) Questions must be naturally and clearly expressed, unambiguous, and free of intentional traps.
(2) Answers must be unique or easily falsifiable (e.g., providing a three-letter English word).

The category-specific simplicity criteria are as follows.
\textbf{\textsc{Rsn}}: Limited to problems solvable with minimal reasoning or intuition.
\textbf{\textsc{Kno}}: Restricted to common knowledge with unique, verifiable answers from sources like Wikipedia.
\textbf{\textsc{If}}: Involve straightforward instructions without strict formatting requirements.
\textbf{\textsc{Ana}}: Limited to questions whose answers can be directly inferred from the prompt, such as binary classification.
These constraints ensure all questions remain straightforward for human respondents.

\subsubsection{A Posteriori Simplicity Verification}

Due to the biases existing between language models and humans~\cite{gallegos2024bias}, questions that are simple for humans may be difficult for LLMs. 
Therefore, we introduce additional posteriori verification to ensure that questions are simple enough to be correctly and robustly answered by smaller LLMs from different families. 
%

\subsection{Construction Workflow}

\noindent \textbf{Subcategory Preparation.}
To ensure diversity, we refer to the subcategories included in existing benchmarks (e.g., MMLU, IFEval, and GSM8K) and evaluation surveys~\cite{chang2024survey} to select, merge, or design subcategories for S1-bench, ensuring that each meets the simplicity requirements.
The definition and example question for each subcategory can be found in Appendix~\ref{app:subcategories}.

\noindent \textbf{Implementation of A Priori Simplicity.}
\textbf{First,} we use two data \textcolor{gray}{\textbf{\textit{generators}}}\footnote{We select Claude-3.7-Sonnet and Qwen2.5-72B-Instruct.} to create 100 initial bilingual question-answer pairs for each candidate subcategory. 
The data generation prompt explicitly incorporates the subcategory definitions, along with both the general and category-specific simplicity criteria, while also aiming to ensure diversity in the generated questions.
\textbf{Second,} these question–answer pairs are then independently evaluated by three annotators and two quality \textcolor{gray}{\textbf{\textit{discriminators}}}\footnote{We select GPT-4o and DeepSeek-V3-241226.} according to the general and category-specific simplicity criteria, resulting in five evaluation outcomes per pair.
The three annotators are experienced graduate students familiar with LLMs and well-acquainted with the goals of S1-Bench.
\textbf{Finally,} based on these evaluation outcomes, three annotators discuss and collectively decide whether to retain, modify, or discard each question (Details in Appendix~\ref{app:human}).

\noindent \textbf{Implementation of A Posteriori Simplicity.}
\textbf{First}, each question obtained from the previous stage is input into the small LLM \textcolor{gray}{\textbf{\textit{validators}}}\footnote{We select four small LLMs: Qwen2.5-7B, Llama3.1-8B, Mistral-8B, and Gemma2-9B. The full model IDs are detailed in Appendix Table~\ref{appendix:small_llm}.} with 7\textasciitilde9 B parameters. For each question, we sample 10 answers at three different temperature settings (0, 0.2, and 0.4), resulting in a total of 30 responses per question.
These responses are then individually evaluated for correctness using GPT-4o.
\textbf{Second}, if all 30 sampled responses are correct, the question is accepted into S1-Bench. Otherwise, the question is returned to the \textcolor{gray}{\textbf{\textit{generators}}}, where a difficulty-reduction prompt is applied to simplify it.
The simplified questions then undergoes the same subsequent process.
\textbf{Finally}, questions fail to meet the full-accuracy criterion (i.e., 30 out of 30 correct) after three rounds of difficulty reduction are excluded from the workflow.

%
The final S1-Bench comprises questions validated by both human a priori constraints and LLM a posteriori verification.
All prompts are in Appendix~\ref{app:s1-cons}.

\input{tables/st_bench_other_llm_acc}

\subsection{Benchmark Statistics}
S1-Bench comprises 422 question-answer pairs across four major categories and 28 subcategories, balanced with 220 English and 202 Chinese questions, with questions averaging 14.46 tokens.
To ensure that the a posteriori verification process does not introduce simplicity only tailored to the small LLM validator, we evaluate S1-Bench on five additional LLMs and on Qwen3 Family with reasoning modes disabled.
As shown in Table~\ref{table:other_llm}, even the 1.7B model achieves over 98\% accuracy at temperature 0.

%% file: tables/st_bench_other_llm_acc.tex
\begin{table}[!t]
\renewcommand{\arraystretch}{1.0}
\setlength\tabcolsep{6.5pt}
\footnotesize
    \centering
    \begin{tabular}{l | ccc | c}
    \toprule
    Model & t=0.0 & t=0.2 & t=0.4 & Tokens \\
    \midrule
    Gemma2-9B    & 100.00 & 100.00 & 100.00 & 38.77 \\
    Llama3.1-8B  & 100.00 & 100.00 & 100.00 & 42.00 \\
    Mistral-8B   & 100.00 & 100.00 & 100.00 & 44.38 \\
    Qwen2.5-7B   & 100.00 & 100.00 & 100.00 & 42.81 \\
    \midrule
    DeepSeek-v3  & 100.00 & 100.00 & 100.00 & 79.53 \\
    Llama3.3-70B & 100.00 & 99.76 & 99.76 & 53.71 \\
    Qwen2.5-14B  & 99.74 & 99.76 & 99.76 & 40.00 \\
    Qwen2.5-32B  & 99.98 & 99.98 & 99.98 & 43.17 \\
    Qwen2.5-72B  & 100.00 & 100.00 & 100.00 & 44.61 \\
    Qwen3-32B (w/o think) & 100.00 & 100.00 & 100.00 &  103.30 \\
    Qwen3-14B (w/o think) & 100.00 & 100.00 & 100.00 &  86.35 \\
    Qwen3-8B (w/o think) & 100.00 & 100.00 & 99.76 &  90.54 \\
    Qwen3-1.7B (w/o think) & 98.10 & 97.16 & 95.73 &  114.32 \\

    \bottomrule    
    \end{tabular}
    \caption{Average accuracy (acc@k) and response token count of different LLMs, each sampled 10 times at three temperature settings.}
    \label{table:other_llm}
\end{table}


%% file: sections/experiment.tex
\section{Experiment Setting}
\subsection{Baseline Models and Configurations}
We evaluated 28 different LRMs, which are explicitly trained to first respond with a \textbf{\textit{thinking process}}, and then generate a \textbf{\textit{final answer}}.
%
%
The evaluated LRMs include both open-source models—such as DeepSeek~\cite{guo2025deepseek}, Qwen~\cite{yang2025qwen3}, Nemotron~\cite{NVIDIA}, Light-R1~\cite{wen2025light}, and s1.1~\cite{muennighoff2025s1}—and closed-source models including claude-sonnet-4-20250514, claude-3-7-sonnet-20250219, gemini-2.5-flash-preview-09-25, o4-mini, and Hunyuan-T1~\cite{Hunyuan}, covering parameter scales from 1.5B to 671B.\footnote{Model details are provided in Appendix Table~\ref{tab:model-settings}.}
For each model, we adopt a temperature of $t=0.6$, top-$p=0.95$, and a sampling size of $k=5$, with the maximum generation length set to 10,000 tokens. As a supplementary exploration, greedy sampling results with temperature t=0 are also provided in the  {Appendix~\ref{app:greedy_results}}.

\input{tables/exp_main_tem_high}

\subsection{Evaluation Metrics}

\paragraph{Format Metrics.} 
We measure format accuracy (\textbf{\textit{f-acc}}) as the proportion of responses that follow the required output format, averaged over five runs. A response is considered valid if an end-of-thinking marker (e.g., \texttt{</think>}) correctly separates the reasoning process from a non-empty final answer.
\paragraph{Efficiency Metrics.}

We report the average response length in tokens (\textbf{\textit{Tokens}}), excluding samples with endless thinking, with all token counts computed using the Qwen2.5 tokenizer to ensure fair comparison.

\paragraph{Accuracy Metrics.}

We calculate accuracy metrics only for responses with correct formatting. Following the LLM-as-judge setting~\cite{zheng2023judging,zhang2023widerdeeperllmnetworks}, we use GPT-4o as the evaluator to assess the \textbf{\textit{final answer}} part. 
We use two metrics: \textit{\textbf{pass@1}} and \textit{\textbf{acc@k}}.
\textit{\textbf{pass@1}} follows the definition in DeepSeek-R1~\cite{guo2025deepseek}, computing the average accuracy over k samples, while \textit{\textbf{acc@k}} is the percentage of questions where all k answers are correct.
Both metrics use k=5.
Notably, f-acc represents the upper bound for pass@1 and acc@k under the formatting requirement.
All details regarding formatting and evaluation can be found in the  {Appendix~\ref{app:B}}.

\input{tables/cat_acc}

\section{Main Experiment}

In this section, we conduct a comprehensive evaluation of LRMs on \bench from three perspectives: accuracy (Table~\ref{tab:main-results}), efficiency (Table~\ref{tab:4-1-tokens-count}), and the impact of RL-based efficient reasoning algorithms (Table~\ref{tab:rl_methods}).

\subsection{Accuracy Analysis}

We analyze the accuracy-related results, and as shown in Table~\ref{tab:main-results}, we obtain the following observations:

\paragraph{Although the vast majority of larger-scale or closed-source models can maintain the same level of accuracy as small LLM-based validators (often exceeding 99\% accuracy), some smaller LRMs suffer from severe performance degradation.}
For example, DS-R1-1.5B and Nemotron-8B achieve acc@k values only slightly above 50\%.
This behavior directly contradicts the original goal of LRMs:
scaling thinking length can actually undermine accuracy on System~1 problems.
Moreover, many LRMs exhibit limited robustness, with pronounced gaps between pass@1 and acc@k consistently observed across models, including DS-R1-1.5B ($\downarrow$26.97\%), L-R1-32B ($\downarrow$15.12\%), DS-R1-7B ($\downarrow$9.43\%), L-R1-7B-DS ($\downarrow$10.99\%), Nemotron-8B ($\downarrow$20.81\%), and s1.1-7B ($\downarrow$24.60\%).
Finally, certain models suffer from output format misalignment, as evidenced by format correctness below 90\% for Nemotron-8B and s1.1-7B.

\input{figure/exp_final_answer_cot_correct}

\input{tables/exp_eff_token_count}
\input{tables/exp_rl_methods}

\paragraph{Accuracy degradation is particularly pronounced on instruction-following and knowledge-based questions.}
We conduct an question type analysis on models whose pass@1 falls below 90\%.
As shown in Table~\ref{tab:cat_acc}, a clear performance gap emerges across categories: instruction-following and knowledge questions reach average accuracies of only 76.7\% and 78.6\%, respectively, versus 88.1\% on reasoning problems.
This pattern suggests that long-chain reasoning does not consistently improve model behavior and may instead weaken instruction-following ability and factual reliability.

\paragraph{We further analyze the error cases and types in the responses.} Specifically, we use DeepSeek-v3 to categorize LRM responses into four cases and compute their proportions.
(1) Final answer correct; thinking process entirely accurate.
(2) Final answer correct; thinking process contains intermediate errors.
(3) Final answer incorrect; correct answer mentioned in thinking process.
(4) Final answer incorrect; correct answer never mentioned in thinking process.
The classification details are in  {Appendix~\ref{app:C}}; results are shown in Figure~\ref{fig:cot_cat_bar}. 
Key findings include:
\textbf{(1) Lower-accuracy LRMs tend to produce less reliable reasoning chains; even when they arrive at the correct final answer, their intermediate steps often contain errors (light green).}
LRMs with high accuracy (e.g., DS-R1) show almost no flawed reasoning steps, whereas those with lower accuracy (e.g., DS-R1-1.5B) often generate incorrect intermediate conclusions, further indicating that they lack robust reasoning ability.
\textbf{(2) Although LRMs sometimes mention the correct answer during reasoning, they may deviate and ultimately produce incorrect final answers (light red).}
We present representative error cases in the {Appendix~\ref{app:error_cases}}. For example, in one case, the LRM initially arrived at the correct answer but undermined it through excessive verification. In another case, the LRM directly denied the correct answer that appeared during reasoning.

\subsection{Efficiency Analysis}

As shown in Table~\ref{tab:main-results}, we present the average response tokens for each problem category in S1-Bench sampled five times (2.11k sample points per model). We observe that: \textbf{(1) all LRMs exhibit overthinking on System 1 problems}.
While closed-source LRMs show higher efficiency, they still consume over three times more tokens. Open-source LRMs are considerably less efficient, with s1.1 and Light-R1 family models exhibiting severe overthinking to achieve better performance on complex problems.
\textbf{(2) LRMs within the same family show similar efficiency without demonstrating clear scaling phenomena}.
Across DS, Qwen3, and Nemotron, response lengths cluster by model family rather than by size, showing no clear correlation between model scale and thinking length. This suggests that training methods and data composition may be more critical factors in determining System 1 thinking effective.

\subsubsection{Analysis across Question Types}

We further present the average response tokens by fine-grained categories and languages, as shown in Table~\ref{tab:4-1-tokens-count}. 
The main findings are as follows:
\textbf{(1) LRMs exhibit a substantial increase in response length across all four major categories, 28 subcategories, and two languages.}
As shown in Table~\ref{tab:4-1-tokens-count}, for each of the four major categories, the average response length of LRMs exceeds that of LLMs by more than a factor of ten.
Response lengths also increase significantly across all subcategories (see Appendix~\ref{app: subcategory_hot}).
This suggests that while LRMs are primarily trained on reasoning data to produce long CoT style responses, this stylistic pattern generalizes well across a wide range of question types.
%
\textbf{(2) LRMs exhibit the most significant increase in tokens for instruction following questions and tend to over-explore when the solution space is vast.}
As shown in Table~\ref{tab:4-1-tokens-count}, although small LLMs provide the most concise responses to instruction following questions, LRMs generate dramatically longer outputs—82.1 $\times $ longer in English and 49.9 $\times $ longer in Chinese than small LLMs.
To investigate the cause, we further analyze the subcategories of instruction following questions.
As shown in {Appendix~\ref{app: subcategory_hot}}, average tokens is notably longer in the subcategories of \textit{length constraints}, \textit{character constraints}, and \textit{sentence constraints}.
These three question types share a similar characteristic: their correctness is easy to verify, but the solution space is vast.
We find that, although the model quickly identifies a correct answer, it becomes trapped in the search space, continually exploring alternatives and failing to stop in time.

\input{figure/exp_first_soluation_number}

\subsubsection{Thinking Solution Analysis}

To better understand the causes of inefficiency in LRMs on \bench, we analyze the solution rounds of their thinking processes
We first use DeepSeek-v3\footnote{See Appendix~\ref{app:robustness} for robustness experiments on automated analysis.} to segment each thinking process into several solutions, each defined as a point at which LRMs explicitly arrives at a conclusion that matches the correct answer. %
We then compute the average token counts in the first solution.
The detailed experimental setup is provided in Appendix~\ref{app: solution_analysis}. Our analysis reveals the following:
\textbf{(1) The token consumed in the first solution of LRMs significantly exceeds that of validator LLMs}.
As shown in Figure~\ref{fig:tokens_bar}~(a),
this suggests that LRMs may involve unnecessary reasoning steps in first solution, which could be one of the reasons for their inefficiency.
\textbf{(2) The primary reason for efficiency gaps between LRMs lies in the number of redundant solution rounds they generate, rather than the token cost in the initial round.}
As shown in Figure~\ref{fig:tokens_bar}~(a), although total thinking token counts vary widely across LRMs, their token counts in the initial round are similar and only account for a small fraction of the total.
Figure~\ref{fig:tokens_bar}~(b) further shows the distribution of solution rounds on \bench, revealing that LRMs with longer thinking processes tend to generate more solution round, and this redundancy greatly increases computational cost.
Furthermore, further experiments reveal that the redundancy in the reasoning process gradually increases over time. 
Furthermore, the similarity across solution rounds progressively increases as reasoning proceeds, suggesting that later rounds contribute increasingly redundant information.
Detailed results can be found in Appendix~\ref{app: thinking_redundancy}.

\subsection{Efficient Reasoning Algorithms Analysis}

Recently, several methods have been proposed to mitigate overthinking in LRMs based on reinforcement learning (RL).
We evaluate representative approaches on \bench, including
(1) RL with length-based penalties (e.g., Shorterbetter~\cite{yi2025shorterbetter}, Laser-D~\cite{liu2025learnreasonefficientlyadaptive}, and TLMRE~\cite{arora2025training}), and
(2) adaptive mode selection (e.g., Adaptthink~\cite{zhang2025adaptthink} and Autothink~\cite{tu2025learning}).
Table~\ref{tab:rl_methods} reports results for models fine-tuned from DS-R1-1.5B/7B, leading to the following observations.
\textbf{(1) Token reduction on challenging benchmarks does not necessarily translate to similar gains on \bench.}
For instance, Laser-D-1.5B and AdaptThink-1.5B reduce reasoning tokens by nearly 50\% on AIME2024, but achieve only about 25\% reduction on S1-Bench, indicating limited generalization of these methods to out-of-distribution problems.
\textbf{(2) Some methods, such as ShorterBetter and AutoThink, achieve substantial compression on \bench at the cost of severe accuracy degradation.}
Specifically, ShorterBetter-1.5B attains only 18.9\% pass@1 on AIME24, while AutoThink-1.5B achieves merely 60.1\% accuracy on \bench, suggesting difficulty in balancing reasoning efficiency with answer correctness.

%% file: tables/exp_main_tem_high.tex
\begin{table}[!t]
\renewcommand{\arraystretch}{1.0}
\setlength\tabcolsep{5pt}
\footnotesize
    \centering
    \begin{tabular}{l| c | cc | c c }
    \toprule

    Model ID & size & pass@1$\uparrow$ & acc@k$\uparrow$ &  f-acc$\uparrow$   &   Tokens $\downarrow$  \\
    \midrule

Validator LLMs & 7-9B & 100.00 & 100.00 & -- &  42.00 \\ 
\midrule
\rowcolor{gray!20} \multicolumn{6}{c}{\textit{close-source LRMs}} \\
claude-sonnet-4 & -- & \textcolor{teal}{\textbf{100.00}} &  \textcolor{teal}{\textbf{100.00}} &    \textcolor{teal}{\textbf{100.00}} & \textcolor{teal}{166.32} \\
claude-3.7-sonnet & -- & \textcolor{teal}{99.95} &  \textcolor{teal}{99.76} &   \textcolor{teal}{\textbf{100.00}} & 178.65 \\
gemini-2.5-flash & -- & 98.48 &  98.01 &   \textcolor{teal}{\textbf{100.00}} & 309.03 \\
o4-mini & -- & 99.53 &  99.53  &   \textcolor{teal}{\textbf{100.00}} & \textcolor{teal}{\textbf{129.99}} \\
Hunyuan-T1 & -- & 99.91 & 99.53 & \textcolor{teal}{\textbf{100.00}} & 542.31 \\

\midrule
\rowcolor{gray!20} \multicolumn{6}{c}{\textit{open-source LRMs}} \\

QwQ-32B & 32B & \textcolor{teal}{\textbf{100.00}} & \textcolor{teal}{\textbf{100.00}} & \textcolor{teal}{\textbf{100.00}} & 720.10 \\
Qwen3-A22B & 235B & 99.91 & \textcolor{teal}{99.76} & \textcolor{teal}{\textbf{100.00}} & 701.65 \\
Qwen3-A3B & 30B & \textcolor{teal}{99.95} & \textcolor{teal}{99.76} & \textcolor{teal}{\textbf{100.00}} & 638.40 \\
Qwen3-32B & 32B & 99.91 & 99.53 & 99.91 & 668.69 \\
Qwen3-14B & 14B & \textcolor{teal}{99.95} & \textcolor{teal}{99.76} & \textcolor{teal}{99.95} & 582.99 \\
Qwen3-8B & 8B & \textcolor{teal}{99.95} & \textcolor{teal}{99.76} & \textcolor{teal}{99.95} & 657.76 \\
Qwen3-1.7B & 1.7B & 99.34 & 97.39 & 99.81 & 595.90 \\

DS-R1 & 671B & \textcolor{teal}{\textbf{100.00}} & \textcolor{teal}{\textbf{100.00}} & \textcolor{teal}{\textbf{100.00}} & 646.40 \\
DS-R1-70B & 70B & 99.38 & 96.92 & 99.91 & 453.81 \\
DS-R1-32B & 32B & 99.72 & 98.82 & \textcolor{teal}{\textbf{100.00}} & 429.91 \\
DS-R1-14B & 14B & 99.57 & 97.87 & \textcolor{teal}{\textbf{100.00}} & 475.46 \\
DS-R1-8B & 8B & 97.39 & 97.16 & 99.53 & 452.11 \\
DS-R1-7B & 7B & 95.21 & 85.78 & 99.24 & 454.55 \\
DS-R1-1.5B & 1.5B & \textcolor{burgundy}{81.47} & \textcolor{burgundy}{\textbf{54.50}} & 97.58 & 489.54 \\

Nemotron-49B & 49B & 99.15 & 97.39 & \textcolor{teal}{\textbf{100.00}} & \textcolor{teal}{\textbf{362.54}} \\
Nemotron-8B & 8B & \textcolor{burgundy}{\textbf{79.81}} & \textcolor{burgundy}{59.00} & \textcolor{burgundy}{\textbf{84.31}} & \textcolor{teal}{372.57} \\

L-R1-32B & 32B & 94.74 & 79.62 & 95.07 & \textcolor{burgundy}{\textbf{1095.36}} \\
L-R1-32B-DS & 32B & 99.57 & 98.10 & 99.81 & 524.12 \\
L-R1-14B-DS & 14B & 99.05 & 95.97 & 99.19 & 693.19 \\
L-R1-7B-DS & 7B & 94.64 & 83.65 & 99.67 & 496.47 \\

s1.1-32B & 32B & 99.48 & 98.10 & 99.53 & \textcolor{burgundy}{998.00} \\
s1.1-14B & 14B & 97.25 & 91.94 & 97.39 & 839.86 \\
s1.1-7B & 7B & 88.58 & 63.98 & \textcolor{burgundy}{88.96} & 711.49 \\

    \bottomrule
    
    \end{tabular}
    \caption{Accuracy results on S1-Bench, sorted by model family. \textcolor{teal}{\textbf{Bold teal}} denotes the best performance, \textcolor{teal}{teal} the second best, \textcolor{burgundy}{\textbf{bold burgundy}} the worst, and \textcolor{burgundy}{burgundy} the second worst.}
    \label{tab:main-results}
\end{table}

%% file: tables/cat_acc.tex
\begin{table}[t]
\renewcommand{\arraystretch}{0.9}
\centering
\footnotesize
\begin{tabular}{lcccccc}
\toprule
Model & pass@1 &  RSN & KNO & IF & ANA \\
\midrule
DS-R1-1.5B     & 81.5 &  86.3 & 75.2 & 76.8 & 84.4 \\
Nemotron-8B    & 79.8 &  84.8 & 71.3 & 76.2 & 83.7 \\
s1.1-7B        & 88.6 &  93.1 & 83.7 & 82.9 & 91.4 \\
Average        & 83.3 &  88.1 & 76.7 & 78.6 & 86.5  \\
\bottomrule
\end{tabular}
\caption{Pass@1 performance across different question types on S1-Bench.
Only models with pass@1 below 90\% on \bench are reported.}
\label{tab:cat_acc}
\end{table}

%% file: figure/exp_final_answer_cot_correct.tex
\begin{figure}[!t]  
\centering  
\includegraphics[width=8cm]{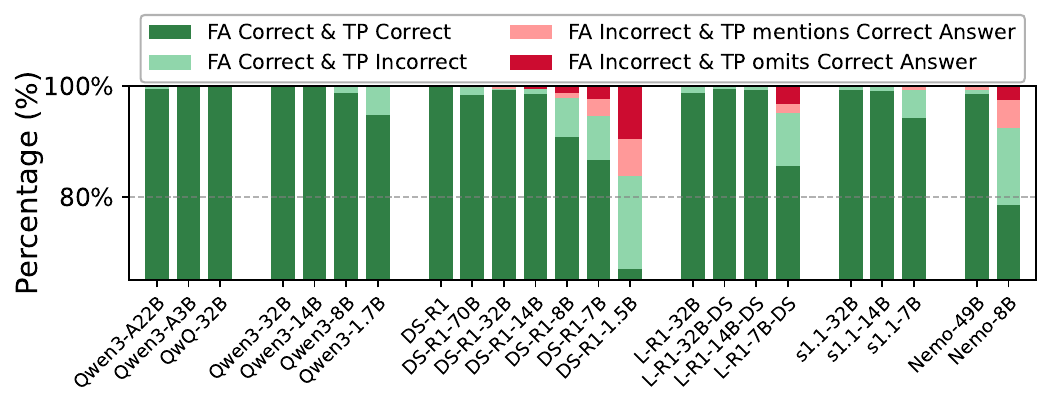}
\vspace{-0.5em}
\caption{Distribution of the thinking process across four categories. FA and TP refer to \textit{\underline{F}inal \underline{A}nswer} and \textit{\underline{T}hinking \underline{P}rocess}, respectively. Green bars indicate cases where the final answer is correct, while red bars indicate cases where it is incorrect.}
\label{fig:cot_cat_bar}
\end{figure}

%% file: tables/exp_eff_token_count.tex
\begin{table*}[!t]
\renewcommand{\arraystretch}{0.9}
\setlength\tabcolsep{5pt}
\footnotesize
    \centering
    \begin{tabular}{l | c | ccccc| ccccc | c }
    \toprule
    \multirow{2}{*}{\raisebox{-0.2\height}{Model ID}} & \multirow{2}{*}{\raisebox{-0.2\height}{size}}  & \multicolumn{5}{c|}{\textbf{S1-Bench-$\mathtt{EN}$}} & \multicolumn{5}{c|}{\textbf{S1-Bench-$\mathtt{ZH}$}} & \multirow{2}{*}{\raisebox{-0.2\height}{\textbf{Avg}}} \\
    
    & & \textsc{Rsn} & \textsc{Kno} & \textsc{If} & \textsc{Ana} & \textbf{Avg}  & \textsc{Rsn} & \textsc{Kno} & \textsc{If} & \textsc{Ana} & \textbf{Avg} & \\
    \midrule
    
Gemma2-9B & 9B & 74.8 & 29.4 & 5.3 & 52.4 & 45.9 & 51.6 & 19.8 & 7.5 & 35.1 & 31.0 & 38.8 \\
Llama3.1-8B & 8B & 91.0 & 35.4 & 12.4 & 61.9 & 56.0 & 44.0 & 28.3 & 15.2 & 18.7 & 26.7 & 42.0 \\
Qwen2.5-7B & 7B & 65.5 & 46.3 & 6.4 & 49.6 & 46.5 & 50.5 & 46.6 & 9.8 & 36.9 & 38.8 & 42.8 \\
Mistral-8B & 8B & 67.2 & 55.5 & 8.6 & 50.1 & 49.6 & 47.3 & 56.1 & 14.8 & 29.7 & 38.7 & 44.4 \\
\midrule
Column Avg & - & 74.6 & 41.6 & 8.2 & 53.5 & 49.5 & 48.3 & 37.7 & 11.8 & 30.1 & 33.8 & 42.0 \\
    
    \midrule

o4-mini & -- & \textcolor{teal}{\textbf{180.5}} & \textcolor{teal}{\textbf{114.2}} & \textcolor{teal}{\textbf{63.1}} & \textcolor{teal}{\textbf{181.1}} & \textcolor{teal}{\textbf{147.2}} & \textcolor{teal}{\textbf{148.1}} & \textcolor{teal}{\textbf{105.9}} & \textcolor{teal}{\textbf{33.5}} & \textcolor{teal}{122.0} & \textcolor{teal}{\textbf{111.2}} & \textcolor{teal}{\textbf{130.0}} \\
claude-sonnet-4 & -- & \textcolor{teal}{188.5} & \textcolor{teal}{127.5} & \textcolor{teal}{66.4} & 225.9 & \textcolor{teal}{168.2} & 180.5 & \textcolor{teal}{149.5} & \textcolor{teal}{65.6} & 204.0 & \textcolor{teal}{164.3} & \textcolor{teal}{166.3} \\
claude-3.7-sonnet & -- & 190.8 & 161.6 & 89.8 & \textcolor{teal}{222.4} & 179.2 & \textcolor{teal}{180.0} & 166.9 & 100.3 & 216.3 & 178.1 & 178.7 \\
gemini-2.5-flash & -- & 295.1 & 141.8 & 122.2 & 398.6 & 268.1 & 383.0 & 220.0 & 271.3 & 465.3 & 353.6 & 309.0 \\
Nemotron-49B & 49B & 599.7 & 587.6 & 396.5 & 526.1 & 540.4 & 232.9 & 157.3 & 235.5 & \textcolor{teal}{\textbf{107.8}} & 168.8 & 362.5 \\
Nemotron-8B & 8B & 561.0 & 585.1 & 458.0 & 303.1 & 462.6 & 369.5 & 326.0 & 288.1 & 166.7 & 273.5 & 372.6 \\
DS-R1-32B & 32B & 421.8 & 504.4 & 414.7 & 521.1 & 473.7 & 362.2 & 385.6 & 343.1 & 408.8 & 382.2 & 429.9 \\
DS-R1-8B & 8B & 472.2 & 528.9 & 530.7 & 462.7 & 491.2 & 521.9 & 404.4 & 266.2 & 395.5 & 409.4 & 452.1 \\
DS-R1-70B & 70B & 464.1 & 501.3 & 378.5 & 536.1 & 484.0 & 450.8 & 450.2 & 328.4 & 416.7 & 420.9 & 453.8 \\
DS-R1-7B & 7B & 447.5 & 623.9 & 353.8 & 510.0 & 495.5 & 446.5 & 463.2 & 339.5 & 373.0 & 409.4 & 454.5 \\
DS-R1-14B & 14B & 503.7 & 674.7 & 367.3 & 494.2 & 519.0 & 452.0 & 465.4 & 375.3 & 405.8 & 428.0 & 475.5 \\
DS-R1-1.5B & 1.5B & 480.8 & 584.7 & 417.4 & 577.2 & 529.1 & 493.0 & 497.4 & 329.8 & 423.1 & 446.0 & 489.5 \\
L-R1-7B-DS & 7B & 568.1 & 667.1 & 501.7 & 566.3 & 580.3 & 444.8 & 454.6 & 344.1 & 366.4 & 405.0 & 496.5 \\
L-R1-32B-DS & 32B & 574.5 & 706.6 & 647.6 & 632.8 & 636.3 & 431.2 & 367.0 & 377.1 & 418.7 & 402.2 & 524.1 \\
Hunyuan-T1 & -- & 561.6 & 693.8 & 380.9 & 435.0 & 521.2 & 676.8 & 553.8 & 505.1 & 523.8 & 565.3 & 542.3 \\
Qwen3-14B & 14B & 700.4 & 639.5 & 286.2 & 575.0 & 579.8 & 730.4 & 557.2 & 403.1 & 586.0 & 586.5 & 583.0 \\
Qwen3-1.7B & 1.7B & 790.4 & 720.6 & 399.9 & 526.2 & 624.6 & 689.8 & 563.6 & 406.4 & 545.9 & 564.7 & 595.9 \\
Qwen3-A3B & 30B & 745.0 & 729.3 & 328.1 & 594.8 & 625.7 & 773.7 & 655.8 & 453.7 & 648.6 & 652.2 & 638.4 \\
DS-R1 & 671B & 786.1 & 723.8 & 711.4 & 529.2 & 672.5 & 727.3 & 638.5 & 607.9 & 533.9 & 617.9 & 646.4 \\
Qwen3-8B & 8B & 853.7 & 753.1 & 394.4 & 629.5 & 683.2 & 749.2 & 623.8 & 459.3 & 624.0 & 630.0 & 657.8 \\
Qwen3-32B & 32B & 805.7 & 774.2 & 356.9 & 645.5 & 674.7 & 780.2 & 695.2 & 446.6 & 645.3 & 662.1 & 668.7 \\
L-R1-14B-DS & 14B & 951.0 & \textcolor{burgundy}{1026.0} & 829.8 & 653.5 & 848.2 & 594.7 & 610.1 & 442.2 & 451.7 & 525.7 & 693.2 \\
Qwen3-A22B & 235B & 925.3 & 864.3 & 487.2 & 605.7 & 734.5 & 803.3 & 713.4 & 487.2 & 611.3 & 665.9 & 701.7 \\
s1.1-7B & 7B & 1039.5 & 840.8 & 1923.2 & 529.4 & 929.9 & 489.6 & 351.3 & 1034.3 & 332.4 & 475.6 & 711.5 \\
QwQ-32B & 32B & 873.3 & 808.1 & 520.8 & 634.7 & 722.4 & 866.9 & 707.3 & 613.3 & \textcolor{burgundy}{\textbf{667.7}} & 717.6 & 720.1 \\
s1.1-14B & 14B & 871.8 & 746.2 & \textcolor{burgundy}{\textbf{2233.1}} & 708.1 & 960.2 & 654.6 & 546.0 & \textcolor{burgundy}{1512.6} & 579.7 & 710.7 & 839.9 \\
s1.1-32B & 32B & \textcolor{burgundy}{1077.9} & 889.7 & \textcolor{burgundy}{2055.4} & \textcolor{burgundy}{781.7} & \textcolor{burgundy}{1081.7} & \textcolor{burgundy}{995.6} & \textcolor{burgundy}{\textbf{765.2}} & \textcolor{burgundy}{\textbf{1634.6}} & \textcolor{burgundy}{666.5} & \textcolor{burgundy}{\textbf{906.5}} & \textcolor{burgundy}{998.0} \\
L-R1-32B & 32B & \textcolor{burgundy}{\textbf{1614.0}} & \textcolor{burgundy}{\textbf{1217.3}} & 1996.9 & \textcolor{burgundy}{\textbf{930.1}} & \textcolor{burgundy}{\textbf{1338.3}} & \textcolor{burgundy}{\textbf{1035.6}} & \textcolor{burgundy}{737.7} & 1240.7 & 610.2 & \textcolor{burgundy}{835.3} & \textcolor{burgundy}{\textbf{1095.4}} \\

\midrule
Column Avg & - & 713.8 & 665.2 & 673.1 & 538.0 & 634.0 & 620.1 & 484.2 & 588.3 & 445.2 & 516.7 & 577.6 \\
Improvement & - & {\textbf{$\times$9.6}} & {\textbf{$\times$16.0}} & {\textbf{$\times$82.1}} & {\textbf{$\times$10.1}} & {\textbf{$\times$12.8}} & {\textbf{$\times$12.8}} & {\textbf{$\times$12.8}} & {\textbf{$\times$49.9}} & {\textbf{$\times$14.8}} & {\textbf{$\times$15.3}} & {\textbf{$\times$13.8}} \\

    \bottomrule

    \end{tabular}
    \caption{Average response tokens in the top-p sampling setting on the S1-bench across two languages and four main categories. \textcolor{teal}{\textbf{Bold teal}} marks best performance, \textcolor{teal}{teal} second best, \textcolor{burgundy}{\textbf{bold burgundy}} worst, and \textcolor{burgundy}{burgundy} second worst. \textbf{Bold} represents the maximum Improvement value for each language.}
    \label{tab:4-1-tokens-count}

\end{table*}

%% file: tables/exp_rl_methods.tex
\begin{table}[!ht]
\renewcommand{\arraystretch}{1.0}
\setlength\tabcolsep{2.1pt}
\footnotesize
    \centering
    \begin{tabular}{l| ccc | ccc | ccc}
    \toprule
    \multirow{2}{*}{\raisebox{-0.4\height}{Model ID}} & \multicolumn{3}{c|}{S1-Bench} & \multicolumn{3}{c|}{MATH500} & \multicolumn{3}{c}{AIME 24} \\
    \cmidrule(lr){2-4} \cmidrule(lr){5-7} \cmidrule(lr){8-10}
    & Tok. & p@1 & \textsc{Rnt} & Tok. & p@1 & \textsc{Rnt} & Tok. & p@1 & \textsc{Rnt} \\
    \midrule
    \rowcolor{gray!20}\multicolumn{10}{c}{\textit{1.5B Model}} \\
    DS-R1-1.5B & 489 & 81.5 & 0.0 & 5534 & 82.1 & 0.0 & 12339 & 28.1 & 0.0 \\
    Shorterbetter-1.5B & 143 & 86.7 & 0.0 & 1131 & 74.8 & 0.0 & 4328 & 18.9 & 0.0 \\
    Laser-D-1.5B & 353 & 85.6 & 0.0 & 1872 & 84.2 & 0.0 & 5750 & 34.2 & 0.0 \\
    TLMRE-1.5B & 185 & 87.2 & 0.0 & 2376 & 84.9 & 0.0 & 9459 & 31.6 & 0.0 \\
    Adaptthink-1.5B & 365 & 84.8 & 23.8 & 1782 & 82.0 & 76.8 & 6670 & 31.0 & 40.4 \\
    Autothink-1.5B & 111 & 60.1 & 91.4 & 2195 & 84.0 & 71.1 & 9514 & 34.6 & 6.4 \\
    
    \midrule
    \rowcolor{gray!20}\multicolumn{10}{c}{\textit{7B Model}} \\
    DS-R1-7B & 454 & 95.2 & 0.0 & 3593 & 92.0 & 0.0 & 10490 & 52.9 & 0.0 \\
    Shorterbetter-7B & 163 & 97.2 & 0.0 & 1210 & 86.6 & 0.0 & 5288 & 53.3 & 0.0 \\
    Laser-D-7B & 351 & 97.3 & 0.0 & 1836 & 92.2 & 0.0 & 5379 & 58.3 & 0.0 \\
    TLMRE-7B & 270 & 97.4 & 0.0 & 2073 & 91.8 & 0.0 & 9410 & 52.3 & 0.0 \\
    Adaptthink-7B & 317 & 97.0 & 26.1 & 1875 & 92.0 & 76.6 & 8051 & 55.6 & 6.3 \\
    Autothink-7B & 66 & 91.8 & 93.3 & 2146 & 91.2 & 12.0 & 8599 & 54.8 & 3.0 \\
    
    \bottomrule
    \end{tabular}
    \caption{Results of RL-based methods for think-token reduction. p@1 denotes pass@1 and \textsc{Rnt} the ratio of no-thinking responses.}
    \label{tab:rl_methods}
\end{table}

%% file: figure/exp_first_soluation_number.tex
\begin{figure}[!t]  
\centering  
\includegraphics[width=7.0cm]{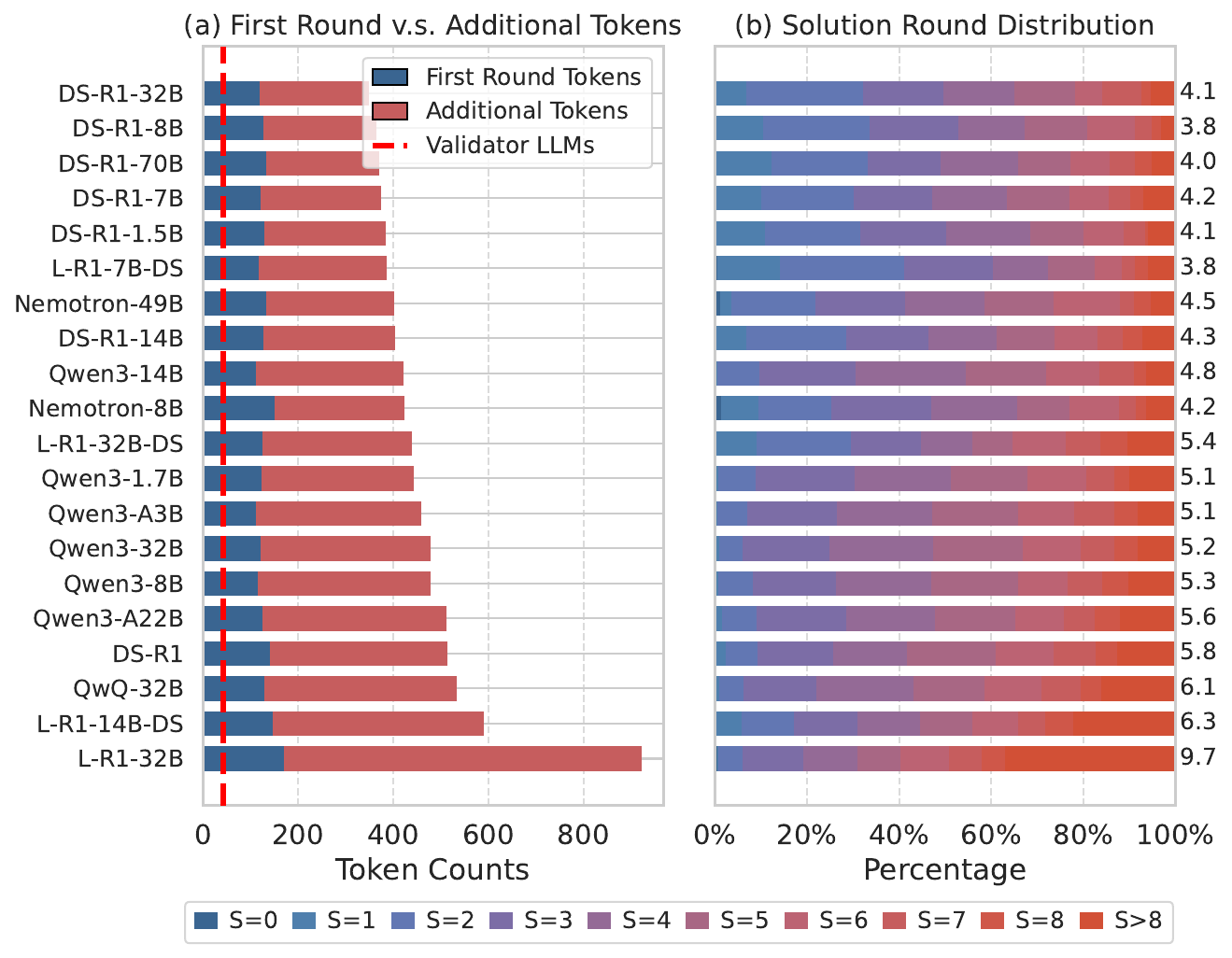}
\caption{(a) Comparison of first round and additional token costs for each LRM. (b) Distribution of solution rounds for each LRM.}
\label{fig:tokens_bar}
\end{figure}

%% file: sections/exp_gut.tex
\section{Early Difficulty Awareness in LRMs}

\input{figure/gut_moment}

We observe that on \bench, LRMs may form an early intuitive judgment of question difficulty before initiating formal reasoning. In some cases, models begin their responses with statements such as “This looks like a simple problem,” prior to any substantive reasoning. We refer to this phenomenon as \textbf{early awareness of question difficulty}. To better understand this behavior, we conduct the following analyses\footnote{See Appendix~\ref{app: error_analysis}, \ref{app: gut}, \ref{app:linear} for experimental setup and details.}.

\paragraph{Exploring Frequency and Stylistic Patterns of Difficulty Awareness.}
We employ GPT-4o to classify the initial portion of model responses—defined as the content before the first “\textbackslash n\textbackslash n”—based on whether and how the model comments on question difficulty. Each response is categorized into one of four types: easy, neutral, difficult, or no comment. Figure~\ref{fig:gut} presents the distribution and corresponding probabilities of these categories across four types of questions.
From these results, we make several observations.
First, all LRMs exhibit early difficulty awareness to varying degrees, with the phenomenon being particularly prominent in the Qwen, DeepSeek, and Light-R1 model families.
Second, LRMs display clear stylistic differences in expressing such awareness. For instance, the Qwen family tends to characterize questions in \bench as easy, whereas DeepSeek-distilled models produce more diverse difficulty-related comments.
Third, this early difficulty awareness is most evident in reasoning-oriented questions, while it appears less frequently in analytical questions.

\paragraph{Exploring the Relationship Between Early Difficulty Awareness, Reasoning Length, and Entropy.}
To investigate whether the early sense of a question as ``easy'' leads to a corresponding reduction in response length, we compare the average response tokens for questions in the easy category versus all samples. The results are shown in Figure~\ref{fig:cat_tokens_diff}.
Except for L-R1-32B, the remaining LRMs do not exhibit a noticeable decrease in response length when questions are perceived as easy. On the contrary, 18 out of 23 LRMs generate longer responses under this condition. 
We further analyze the average token entropy of responses associated with easy judgments, with the results presented in Figure~\ref{fig:entropy}. Interestingly, these responses exhibit higher average entropy, indicating stronger exploratory behavior during generation.
One possible explanation is that models may exhibit self-doubt toward their own early difficulty judgments, leading them to engage in increased exploration and produce longer responses, despite initially perceiving the question as simple.
\textbf{This suggests a discrepancy between the LRM’s initial sense of difficulty and its generative behavior}, the causes and improvements of which warrant further investigation.

\input{figure/cat_tokens_diff}
\input{figure/three_entropy}

\paragraph{Exploring Activation Pattern Differences Across Question Difficulty.}
In contrast to the preceding analyses based on explicit difficulty-related cues in model outputs, we further examine whether question difficulty is reflected at the representation level through systematic differences in model activation patterns.
Specifically, using DS-R1-1.5B and DS-R1-7B as representative examples, we analyze whether the final-layer representation of the last token in the encoded question varies systematically with question difficulty.
To this end, we conduct a probing-based analysis by training a single-layer MLP as a weak linear classifier with limited training data, which serves solely to assess whether hidden representations contain information predictive of question difficulty.
The training set is constructed by selecting one Chinese and one English example from each subcategory of \bench as easy questions, and sampling an equal number of bilingual questions from AIME25 as hard questions, where Chinese examples are obtained via translation from English.
We then evaluate the trained probe on the MATH500 dataset, which naturally contains five difficulty levels. For each level, we compute the proportion of questions classified as difficult. As shown in Table~\ref{tab:difficulty_probe}, this proportion increases consistently with question difficulty, indicating that model representations implicitly encode structured information aligned with problem difficulty.

\begin{table}[t]
\centering
\footnotesize
\begin{tabular}{lccccc}
\toprule
Model & Level 1 & Level 2 & Level 3 & Level 4 & Level 5 \\
\midrule
DS-R1-1.5B & 46.5 & 56.7 & 56.7 & 79.7 & 92.5 \\
DS-R1-7B   & 27.9 & 43.3 & 60.9 & 81.3 & 91.1 \\
\bottomrule
\end{tabular}
\caption{Proportion (\%) of questions classified as difficult across five difficulty levels on the MATH500 dataset.}
\label{tab:difficulty_probe}
\end{table}

%% file: figure/gut_moment.tex
\begin{figure}[!t]  
\centering  
\includegraphics[width=7.5cm]{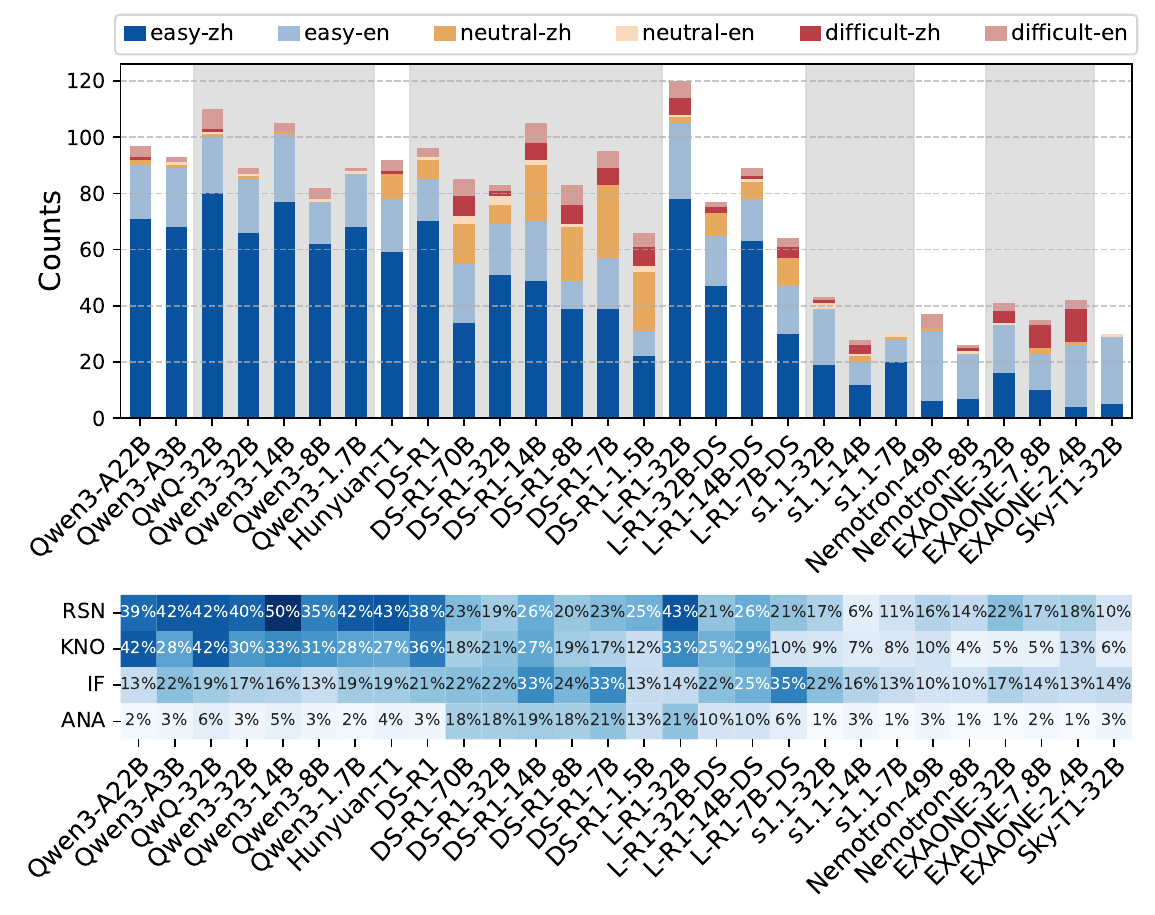}
\vspace{-1em}
\caption{Top: Count of early difficulty awareness across models. Bottom: Probability of early difficulty awareness by question type.}
\label{fig:gut}
\end{figure}

%% file: figure/cat_tokens_diff.tex
\begin{figure}[!t]  
\centering  
\includegraphics[width=7.5cm]{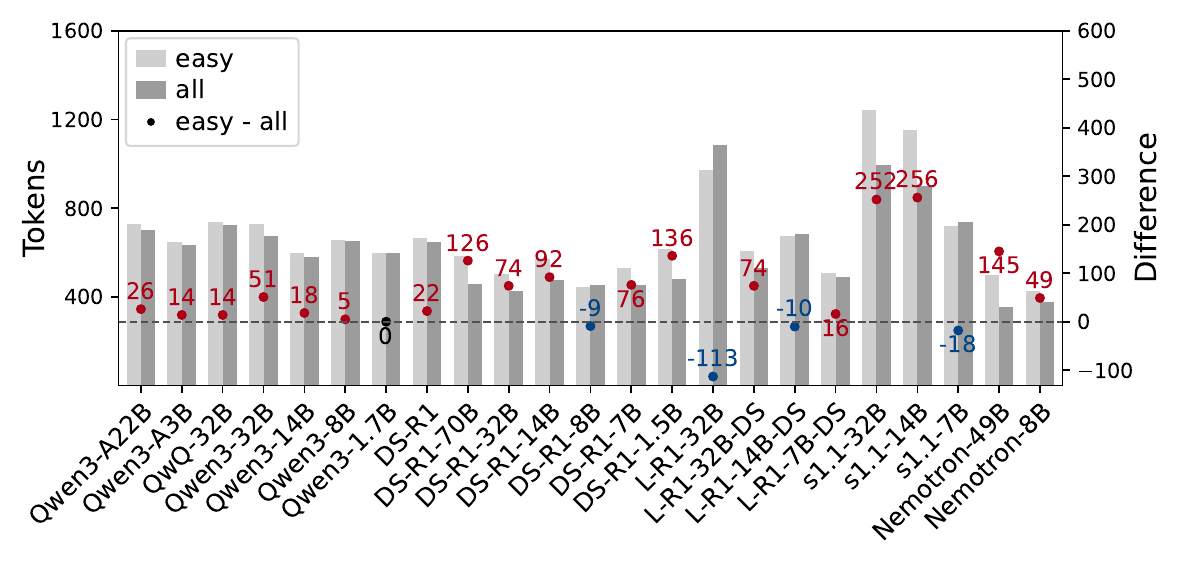}
\vspace{-1.1em}
\caption{Average response tokens in the easy category vs. all samples. Dots show difference: easy minus all.}
\label{fig:cat_tokens_diff}
\end{figure}

%% file: figure/three_entropy.tex
\begin{figure}[!t]  
\centering  
\includegraphics[width=8.5cm]{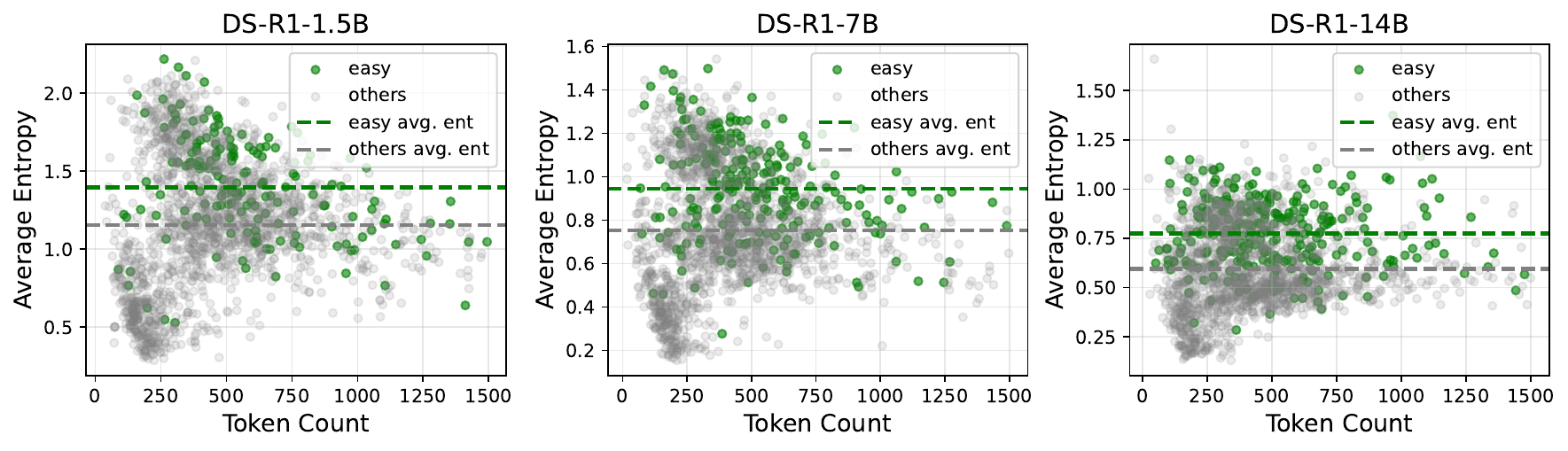}
\vspace{-1.5em}
\caption{Average response entropy of easy and other responses across three LRMs.}
\label{fig:entropy}
\end{figure}

%% file: sections/related-work.tex
\section{Related work}

\subsection{Large Reasoning Models}
Large Reasoning Models (LRMs), characterized by explicitly generating external thinking processes before final answers~\cite{LRM-survey-2,LRM-survey-3}, achieve a paradigm shift from intuitive \sysone thinking to deliberative \systwo reasoning compared to traditional LLMs~\cite{li2025survey,qu2025survey}, thus achieving superior performance on complex tasks.
The development of recent LRMs has largely followed two main approaches: large-scale reinforcement learning (RL) and model distillation.
Models trained via RL~\cite{shape,team2025kimi,attnpo} leverage reward-based optimization to gradually incentivize deliberative reasoning. In contrast, distillation-based LRMs~\cite{openai-o1,min2024imitate,team2025sky,ye2025limo,muennighoff2025s1,zhang2025sotopiaomegadynamicstrategyinjection} acquire such abilities by transferring structured reasoning patterns from advanced teacher models.

\subsection{Limitations of LRMs}

While LRMs have shown significant performance gains through deliberate reasoning, rigid adherence to this overly cautious thinking can introduce new limitations. On one hand, intermediate reasoning steps can cause excessive token generation and unnecessary solving attempts~\cite{chen2024overthink,acc-drop-dna,overthink-kumar}, even leading to redundancy in the hidden layers~\cite{chen2025inner}. On the other hand, LRMs' performance can drop in specific contexts like safety scenarios~\cite{jiang2025safechain}, agents~\cite{expseek} and role-playing~\cite{roleplay}. However, prior studies mainly evaluated LRMs on complex tasks that more suited for deliberative \systwo thinking. Our work examines how deliberative reasoning impacts extremely simple problems better matched to intuition-driven \sysone processing.

%% file: sections/conclusion.tex
\section{Conclusion}

This paper extensively explores the \sysone thinking capabilities of LRMs. We construct S1-Bench, a simple, domain-diverse, and natural benchmark for evaluating \sysone thinking abilities. Our exploration reveals fundamental issues of LRMs on \sysone problems—under-accuracy and over-reasoning—and exposes limitations of existing efficient training methods. We also demonstrate models' inherent perception of \sysone problems from both linguistic and representational perspectives, providing insights for future research.

%% file: sections/acknowledgments.tex
\section*{Acknowledgments}
We would like to thank the anonymous reviewers, the area chairs and program chairs for their valuable comments and efforts.
This work is supported by the National Natural Science Foundation of China (Grant No. 62572465).

%% file: sections/appendix.tex
\section{More Information of \bench Construction}

\subsection{Benchmark Statistics}
\label{app:count}

We survey studies on improving the efficiency of LRMs, as there is potential overlap between these studies and the technical approaches aimed at enhancing \sysone thinking in LRMs. 
Table~\ref{app:count_benchmark} presents the results of our survey. 
We compile the benchmarks used in these studies for evaluation, that are typically used to verify whether models achieve efficiency improvements. 
Benchmarks that appear more than four times include: MATH500 \cite{MATH}, GSM8K \cite{GSM8K}, AIME24/25 \cite{aime}, GPQA \cite{rein2024gpqa}, AMC23 \cite{AMC2023}, MMLU \cite{hendrycksmeasuring}, Olympiad-Bench \cite{he2024olympiadbench}, SVAMP \cite{patel2021nlp}, LiveCodeBench \cite{jainlivecodebench}, and CommonSenseQA \cite{talmor2019commonsenseqa}.

The accuracy shown in Table 1 of the main paper is the average result of the four models, Qwen2.5-7B, Llama3.1-8B, Mistral-8B, and Gemma2-9B, at temperature 0, using GPT-4o as the evaluator.

\subsection{Subcategories in S1-Bench}
\label{app:subcategories}
Figure~\ref{fig:pie} shows the pie chart distribution of 28 subcategories in S1-Bench. For more details on the subcategories, please refer to Table~\ref{tabel:sub_cat_exp_1}, \ref{tabel:sub_cat_exp_2}.
We present the names of all subcategories along with a representative question, making every effort to ensure orthogonality across all categories to improve the diversity of S1-Bench.

\subsection{Prompt for S1-Bench construction}
\label{app:s1-cons}
This section presents the prompts used in the construction of S1-Bench, including the Initial Generation prompt, the Discriminating Generation Quality prompt, and the Reduce Difficulty prompt. See Table~\ref{table:prompt_s1_workflow} for details.

\subsection{Human-involved Construction}
\label{app:human}
All annotators are native Chinese speakers with high English proficiency. The annotators are well-versed in the objectives of benchmark construction and received concentrated training to ensure that all questions have unique answers and are as distinct and non-repetitive as possible. After the benchmark was finalized, all three annotators re-examined every question to ensure data quality, removing all inconsistent questions.

\input{figure/st_bench_pie}
\section{Baseline Models and Evaluation Details}
\label{app:B}
\subsection{Baseline Model Details}
\label{app:model_details}
Table~\ref{appendix:small_llm} presents the abbreviations, IDs, and URLs of LLMs used in this paper. 
Table~\ref{tab:model-settings} displays the abbreviations, IDs, URLs, organizations, training algorithms, and training data volumes of open-source LRMs evaluated in this study.

\subsection{Efficient Reasoning Algorithms Settings}
We evaluate all RL-based methods using publicly released checkpoints from the original authors.
We sample 5 responses per prompt on \bench (max length 10,000), 1 on Math500 (max length 16,384), and 16 on AIME2024 (max length 16,384).
All models use top-p sampling with temperature 0.6 temperature $=$ 0.6 and p$=$0.95.

\subsection{GPT-4o and Human Evaluation}
\label{app:eval_4o}
We use GPT-4o as the evaluator to assess the correctness of the responses.
If a final answer can be isolated, only the final answer is evaluated; otherwise, the entire response is assessed.
The evaluation prompt is provided in Table~\ref{prompt:eval_4o}.

To evaluate the consistency between the GPT-4o judge’s assessments and human judgments, we conduct a comprehensive human evaluation study involving three of the authors. 
All participating authors are proficient in both English and Chinese.
Specifically, we randomly sample 20 question-answer pairs from each model's greedy decoding results, resulting in a dataset of 640 pairs derived from 32 models (including 4 verifier LLMs and 28 LRMs).
The questions, reference answers, and model responses are then presented to three annotators, who independently judge the correctness of each model response. 
The final human evaluation results are determined through majority voting. 
Ultimately, the Cohen’s Kappa between the human evaluators and the GPT-4o judge is calculated to be 0.83, indicating an exceptionally high level of agreement.

\subsection{Accuracy Metrics Details}
\label{app:acc_metric_details}
\textbf{Pass@1:} Followed DeepSeek-R1~\cite{guo2025deepseek}, we calculate pass@1 to assess the percentage of correct responses among the k=5 generations. Specifically, it is defined as:
\begin{equation}
\text{pass@1}=\frac{1}{k}\sum_{i=1}^{k}p_i ,
\end{equation}
where $p_i$ is the correctness of the i-th generation.\\
\textbf{Acc@k:} Since S1-Bench is composed of extremely simple questions, we calculate acc@k. Specifically, acc@k=1 if all k responses are correct and acc@k = 0 otherwise. It is defined as:
\begin{equation}
\text{acc@k} = \prod_{i=1}^{k}p_{i},
\end{equation}

\input{tables/exp_main_tem_zero}

\subsection{Types and Analysis of Format Errors}
\label{app:format_details}

This section introduces a comprehensive taxonomy of format errors and emphasizes the importance of addressing these issues in future research, \textit{particularly in the design of format rewards for reinforcement learning}.
Unlike conventional LLMs, LRMs frequently exhibit format errors. 
These errors are defined by failing to use a unique end thinking marker (e.g., \texttt{</think>}) to separate the thinking process from the final answer. 
Format errors increase the difficulty of distinguishing the thinking process from the final answer and reveal the vulnerability of LRMs in following predefined formats.

To illustrate this phenomenon, we identify 12 distinct types of response formats produced by LRMs, each assigned a unique ID, as shown in Table~\ref{tab:format-error-type}. 
These 12 types are further grouped into three major categories:
\begin{itemize}
    \item Standard-Conforming Responses: These responses meet the expected format by including exactly one end thinking marker (e.g., \texttt{</think>}) to delimit the thinking process from the final answer. Among these, type ID-100 includes a thinking process, while ID-101 omits it. The proportion of such responses is measured using the S-Corr metric.
    
    \item Unreadable Responses: These refer to generation failures, including cases where LRMs produce endlessly thinking content or solely produce end thinking markers. The proportion of all other (i.e., readable) responses is measured using the L-Corr metric.
    
    \item Readable but Malformed Responses: These responses deviate from the standard format yet still contain extractable information. In some cases, the final answer is missing (e.g., ID-200, ID-202, ID-205), and we instead evaluate the correctness of the thinking process. In other cases, multiple (e.g., ID-201, ID-203) or unmatched\footnote{This paper provides a reference collection of unmatched end thinking makers: \texttt{</ think>}, \texttt{</th think>}, \texttt{</ reason>}, \textbackslash nanswer\textbackslash n ,**Final Answer** and \begin{CJK}{UTF8}{gkai}**答案**\end{CJK}.} (e.g., ID-204, ID-206) end thinking markers are generated. In such instances, we treat the content following the last end thinking marker as the final answer for evaluation.
\end{itemize}

Table~\ref{table:error_counts_rates_high_t} and Table~\ref{table:error_counts_rates_zero_t} present the distributions of 12 format types under top-p sampling and greedy sampling, respectively.
we find:
(1) The infinite generation phenomenon is widespread across most LRMs, particularly concentrated in LRMs with fewer than 32B parameters.
%
%
(2) The Nemotron family frequently produces correctly formatted responses without any explicit thinking processes. 
This behavior can be viewed as a mechanism for mitigating over-thinking. 
(3) None of the evaluated LRMs exhibited behaviors classified as ID-205/206.



\input{tables/exp_error_format_types}

\input{figure/exp_cos_sim}

\section{More Experimental Setups \& Results}
\label{app:C}

\subsection{Greedy Sampling Results}
\label{app:greedy_results}
To distinguish between the two settings, we refer to the main text sampling as top-p sampling.
Table~\ref{tab:main-results-temp0} presents the performance of LRMs on S1-Bench under greedy sampling. While overall accuracy improves compared to top-p sampling, issues of inefficiency and accuracy degradation on simple questions remain.

\subsection{Efficiency Analysis across Subcategories.}
\label{app: subcategory_hot}
Figure~\ref{fig:hot} illustrates the average response tokens across the 28 subcategories.
In the heatmap, both models (rows) and subcategories (columns) are ordered in descending order according to their average number of response tokens.

\subsection{Solution Analysis Details}
\label{app: solution_analysis}
For solution analysis, We only use well-formatted thinking processes with correct final answers, as incorrect answers make it unclear whether LRMs are over-reasoning or under-reasoning, and malformed thinking processes cannot be precisely extracted.
The segmentation process is performed by DeepSeek-v3, with prompts detailed in Table~\ref{prompt:segmentation}.
We compute the average token count in the first solution round; if no solution is found, we use the token count of the entire thinking process.

\subsection{Thinking Redundancy Analysis}
\label{app: thinking_redundancy}
We conduct a similarity analysis to analyze how information redundancy in the thinking processes changes as reasoning sequences increase. 
Specifically, we first divide the complete thinking process into k equal-length segments\footnote{We set k=15, changing its value does not affect the conclusions.}.
Then, we encode each segment using the all-MiniLM-L6-v2 model\footnote{https://huggingface.co/sentence-transformers/allMiniLM-L6-v2}.
For each segment, we calculate the cosine similarity with all its preceding segments and use the maximum similarity as a measure of its information redundancy.
As shown in Figure~\ref{fig:cos_sim}, information redundancy increases across all four main categories as reasoning sequences increase. 
%

\input{tables/gut_show}

\input{tables/appendix_human_review}

\input{tables/appendix_llm_details}
\input{tables/appendix_lrm_details}
\input{tables/appendix_count_benchmark}
\input{tables/st_bench_sub_cat_intro}
\input{tables/st_bench_sub_cat_intro_2}
\input{prompt/S1_bench_workflow}

\input{prompt/eval_4o}
\input{figure/exp_hot}
\input{tables/exp_error_format_number_temp}

\input{prompt/segment}
\input{prompt/cot_eval_correct_wrong}

\input{prompt/gut_en}
\input{tables/case1}

\subsection{Error Analysis Details}
\label{app: error_analysis}

In error analysis, we only use well-formatted samples, as malformed thinking processes cannot be precisely extracted.
For samples with correct final answers, we categorize them based on whether the thinking process contains explicit incorrect conclusions in intermediate steps. 
For samples with incorrect final answers, we categorize them based on whether the correct answer is mentioned at least once during reasoning.
We use DeepSeek-v3 for categorization, with prompts provided in Table~\ref{table:prompt_error}.

\subsection{Difficulty Awareness Analysis Details}
\label{app: gut}
We prompt GPT-4o to classify the initial part of model responses (before the first `\textbackslash n\textbackslash n') into four types based on its comment on difficulty: easy, neutral, difficult, and no comment. 
The prompts for english question can be seen in Table~\ref{prompt:gut-en}.
For Chinese queries, we use the translated version of the prompt in Chinese.
In Table~\ref{tabel:gut_show}, we show the most common sentence of all LRMs in each type of early difficulty awareness.

\subsection{Linear Classifier Training Configuration}
\label{app:linear}
We train the linear prob for two epochs using a learning rate of 1e-3 and binary cross-entropy loss. Training completes in less than 10 seconds on a single RTX 3090 GPU.

\subsection{Robustness of Automated Analysis}
\label{app:robustness}
Manual analysis of 28 models across two settings totaling 11.8K responses with an average length of 650 tokens each is impractical, while manually analyzing random subsets introduces bias. We aim to analyze all data to obtain robust conclusions.

When conducting these two sets of experiments, we randomly sampled 200 responses for extensive pre-experiments (different prompts, analysis models, etc.) and conducted manual review (by three authors) on the accuracy of error analysis and difficulty awareness extraction. Specifically, error analysis was modeled as a four-class classification task, and difficulty awareness was modeled as a four-class classification task (easy, neutral, difficult, and no comment). We found that DeepSeek-V3, which excels at instruction following, has dual advantages in both accuracy and cost compared to Qwen2.5-72B-Instruction and GPT-4o. Additionally, we carefully designed different prompts for Chinese and English and tried multiple approaches, as the two languages differ in expression habits and extraction boundaries. The manual review results of our final approach are shown in Table~\ref{tab:manual_review}.

\section{Error Cases}
\label{app:error_cases}
This section presents several error cases observed in LRMs. 
See Tables~\ref{case: 7777},~\ref{case:cat},~\ref{case:Sydney},~\ref{case: Beethoven},~\ref{case: 1.5+3.5}, and~\ref{case:1821}.


%% file: figure/st_bench_pie.tex
\begin{figure}[!h]  
\centering  
\includegraphics[width=7.5cm]{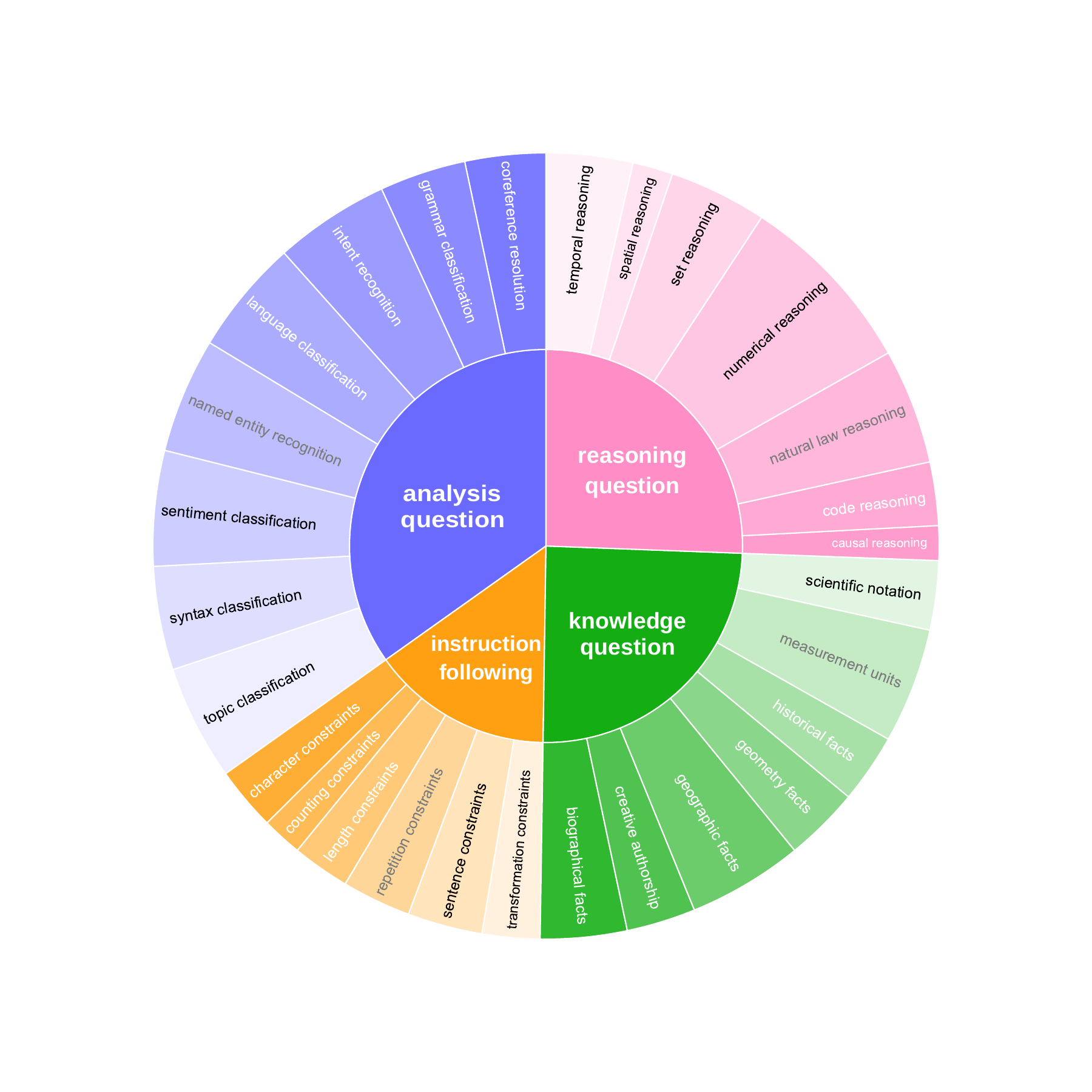}
\caption{S1-Bench Category Display. The inner circle represents four major categories, and the outer circle includes 28 subcategories. Detailed information is presented in~\ref{tabel:sub_cat_exp_1} and~\ref{tabel:sub_cat_exp_2}.}
\label{fig:pie}
\end{figure}

%% file: tables/exp_main_tem_zero.tex
\begin{table}[!t]
\renewcommand{\arraystretch}{1.1}
\setlength\tabcolsep{5pt}
\small
    \centering
    \begin{tabular}{l| c | c | c c }
    \toprule
    Model & size & acc & f-acc $\uparrow$  &  tokens $\downarrow$  \\
    \midrule

Qwen3-235B-A22B & 235B & \textcolor{teal}{\textbf{100.00}} & \textcolor{teal}{\textbf{100.00}} & 702.70 \\
Qwen3-30B-A3B & 30B & \textcolor{teal}{\textbf{100.00}} & \textcolor{teal}{\textbf{100.00}} & 636.35 \\
QwQ-32B & 32B & \textcolor{teal}{\textbf{100.00}} & \textcolor{teal}{\textbf{100.00}} & 750.41 \\
Qwen3-32B & 32B & \textcolor{teal}{99.76} & \textcolor{teal}{99.76} & 673.62 \\
Qwen3-14B & 14B & \textcolor{teal}{99.76} & \textcolor{teal}{\textbf{100.00}} & 597.06 \\
Qwen3-8B & 8B & \textcolor{teal}{99.76} & \textcolor{teal}{\textbf{100.00}} & 649.45 \\
Qwen3-1.7B & 1.7B & 99.53 & \textcolor{teal}{99.76} & 579.01 \\


DS-R1 & 671B & \textcolor{teal}{\textbf{100.00}} & \textcolor{teal}{\textbf{100.00}} & 621.89 \\
DS-R1-70B & 70B & \textcolor{teal}{99.76} & \textcolor{teal}{\textbf{100.00}} & 469.78 \\
DS-R1-32B & 32B & \textcolor{teal}{\textbf{100.00}} & \textcolor{teal}{\textbf{100.00}} & \textcolor{teal}{428.46} \\
DS-R1-14B & 14B & 99.29 & \textcolor{teal}{99.76} & 463.52 \\
DS-R1-8B & 8B & 97.39 & 99.53 & 452.11 \\
DS-R1-7B & 7B & 94.31 & 97.87 & 436.87 \\
DS-R1-1.5B & 1.5B & \textcolor{burgundy}{\textbf{76.54}} & 91.94 & 473.67 \\


Nemotron-49B & 49B & 99.53 & \textcolor{teal}{\textbf{100.00}} & \textcolor{teal}{\textbf{337.94}} \\
Nemotron-8B & 8B & \textcolor{burgundy}{77.73} & \textcolor{burgundy}{\textbf{81.99}} & 446.62 \\

L-R1-32B & 32B & 85.78 & \textcolor{burgundy}{85.78} & \textcolor{burgundy}{\textbf{996.36}} \\
L-R1-32B-DS & 32B & 99.29 & 99.29 & 528.45 \\
L-R1-14B-DS & 14B & 98.82 & 98.82 & 664.28 \\
L-R1-7B-DS & 7B & 92.65 & 98.82 & 514.60 \\

s1.1-32B & 32B & 98.82 & 98.82 & \textcolor{burgundy}{983.38} \\
s1.1-14B & 14B & 95.50 & 95.97 & 786.30 \\
s1.1-7B & 7B & 87.68 & 87.91 & 630.52 \\

    \bottomrule
    
    \end{tabular}
    \caption{Main results in the greedy decoding setting on the S1-Bench, sorted by model family. \textcolor{teal}{\textbf{Bold teal}} marks best performance, \textcolor{teal}{teal} second best, \textcolor{burgundy}{\textbf{bold burgundy}} worst, and \textcolor{burgundy}{burgundy} second worst.}
    \label{tab:main-results-temp0}
\end{table}

%% file: tables/exp_error_format_types.tex
\begin{table}[h!]
\renewcommand{\arraystretch}{1.1}
\setlength\tabcolsep{3.5pt}
\footnotesize
    \centering
    
    \begin{tabular*}{0.48\textwidth}{@{}l | c | @{\extracolsep{\fill}} c c c c c @{}}
    \toprule
    \multirow{2}{*}{\raisebox{-0.2\height}{Format}} & \multirow{2}{*}{\raisebox{-0.2\height}{ID}} & {\raisebox{-0.3\height}{marker}} & {\raisebox{-0.3\height}{marker}} & {\raisebox{-0.3\height}{marker}} & {\raisebox{-0.3\height}{thinking}} & {\raisebox{-0.3\height}{final}} \\
    & & (standard) & (unmatched) & (number) & process &  answer \\
    \midrule
    \multirow{2}{*}{\raisebox{-0.2\height}{Standard}}
     & 100 & $\surd$ & -- & 1 & $\surd$ & $\surd$ \\
     & 101 & $\surd$ & -- & 1 & $\times$ & $\surd$ \\
    \midrule
    \multirow{8}{*}{\makecell[l]{Readable but \\ Malformed}}
     & 200 & $\surd$ & -- & 1 & $\surd$ & $\times$ \\
     & 201 & $\surd$ & -- & $>$1 & $\surd$ & $\surd$ \\
     & 202 & $\surd$ & -- & $>$1 & $\surd$ & $\times$ \\
     & 203 & $\surd$ & -- & $>$1 & $\times$ & $\surd$ \\
     & 204 & $\times$ & $\surd$ & $\ge$1 & $\surd$ & $\surd$ \\
     & 205 & $\times$ & $\surd$ & $\ge$1 & $\surd$ & $\times$ \\
     & 206 & $\times$ & $\surd$ & $\ge$1 & $\times$ & $\surd$ \\
     & 207 & $\times$ & $\times$ & 0 & -- & $\surd$ \\
    \midrule
    \multirow{2}{*}{\raisebox{-0.2\height}{Unreadable}}
    & 300 & $\surd$ & $\surd$ & $\ge$1 & $\times$ & $\times$ \\
    & 301 & $\times$ & $\times$ & 0 & -- & $\infty$  \\
    
    \bottomrule
    
    \end{tabular*}
    \caption{Twelve types of response format.}
    \label{tab:format-error-type}
\end{table}

%% file: figure/exp_cos_sim.tex
\begin{figure*}[!t]  
\centering  
\includegraphics[width=16.5cm]{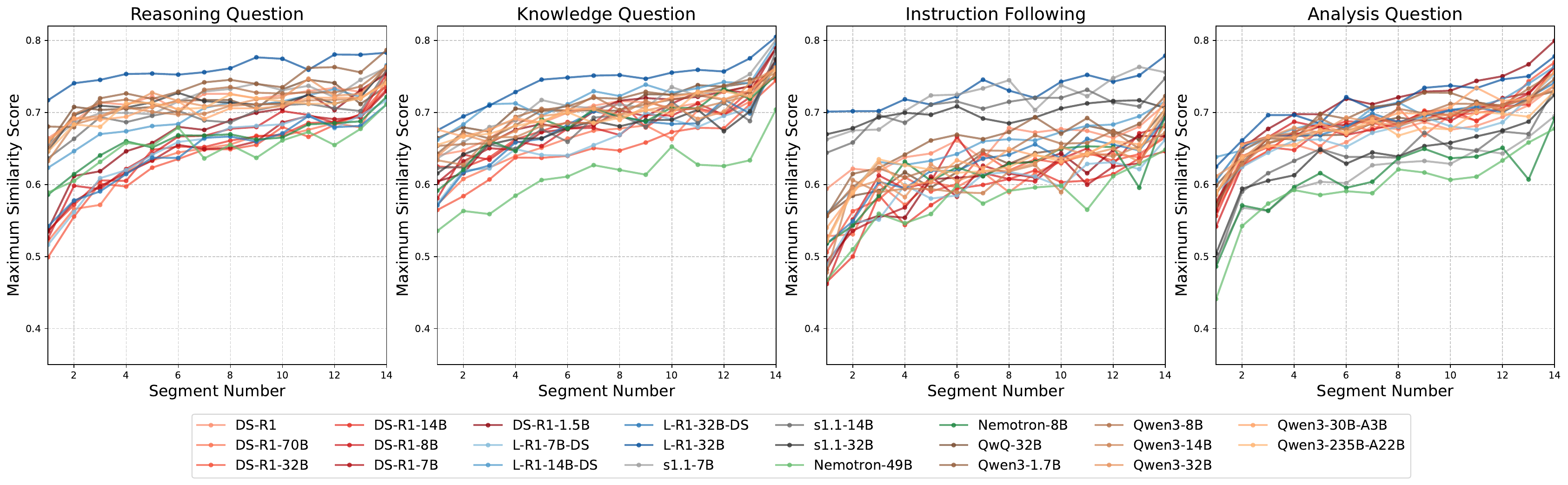}

\caption{Maximum similarity between each segment and all preceding segments for open-source LRMs across four categories.}
\label{fig:cos_sim}
\end{figure*}

%% file: tables/gut_show.tex
\begin{table}[!t]
\renewcommand{\arraystretch}{1.1}
\setlength\tabcolsep{5pt}
\small
    \centering
    \begin{tabular}{c | c | c }
    \toprule
    Type & Sentence & Count \\
    \midrule
    easy-zh & \begin{CJK}{UTF8}{gkai} 这个问题看起来挺简单的 \end{CJK} & 308\\
    easy-en & that seems straightforward & 36\\
    difficult-zh & \begin{CJK}{UTF8}{gkai} 这个问题看起来有点复杂 \end{CJK} & 308\\
    difficult-en & percentages can sometimes be tricky & 7\\
    neutral-zh & \begin{CJK}{UTF8}{gkai} 这个问题看起来好像不难 \end{CJK} & 24\\
    neutral-en & Hmm, interesting & 3\\

    \bottomrule
    \end{tabular}
    \caption{The most common sentence in each type of early difficulty awareness.}
    \label{tabel:gut_show}
\end{table}

%% file: tables/appendix_human_review.tex
\begin{table*}[t]
\centering
\small
\begin{tabular}{lcccccc}
\toprule
\multirow{2}{*}{\raisebox{-0.2\height}{Model}} & avg Acc & Kappa & avg Acc & Kappa & Input Price & Output Price \\
 & (error ana) & (error ana) & (diff awareness) & (diff awareness) & (API) & (API) \\
\midrule
Qwen2.5-72B-Instruction & 90.0\% & 0.51 & 94.5\% & 0.66 & \$0.54 & \$1.62 \\
GPT-4o & 93.0\% & 0.56 & 98.0\% & 0.70 & \$2.50 & \$10.00 \\
DeepSeek-V3-0324 & 90.5\% & 0.63 & 97.5\% & 0.68 & \$0.27 & \$1.10 \\
\bottomrule
\end{tabular}
\caption{Manual review results.}
\label{tab:manual_review}
\end{table*}

%% file: tables/appendix_llm_details.tex
\begin{table*}[!t]
\small
    \centering
    \begin{tabular}{l l l}
    \toprule
    Model & Model ID & URL \\
    \midrule
    Qwen2.5-7B & Qwen2.5-7B-Instruct & https://huggingface.co/Qwen/Qwen2.5-7B-Instruct \\
    Llama3.1-8B & Llama-3.1-8B-Instruct & https://huggingface.co/meta-llama/Llama-3.1-8B-Instruct \\
    Mistral-8B & Ministral-8B-Instruct-2410 & https://huggingface.co/mistralai/Ministral-8B-Instruct-2410 \\
    Gemma2-9B & gemma-2-9b-it & https://huggingface.co/google/gemma-2-9b-it \\
    \midrule
    Qwen2.5-14B  & Qwen2.5-14B-Instruct & https://huggingface.co/Qwen/Qwen2.5-14B-Instruct \\
    Qwen2.5-32B  & Qwen2.5-32B-Instruct & https://huggingface.co/Qwen/Qwen2.5-32B-Instruct \\
    Qwen2.5-72B  & Qwen2.5-72B-Instruct & https://huggingface.co/Qwen/Qwen2.5-72B-Instruct \\
    Llama3.3-70B & Llama-3.3-70B-Instruct  & https://huggingface.co/meta-llama/Llama-3.3-70B-Instruct \\
    DeepSeek-v3  & DeepSeek-V3-0324 & https://huggingface.co/deepseek-ai/DeepSeek-V3-0324 \\
    
    \bottomrule    
    \end{tabular}
    \caption{Mapping of LLM abbreviations and IDs used in this paper, with their open-source URLs.}
    \label{appendix:small_llm}
\end{table*}

%% file: tables/appendix_lrm_details.tex
\begin{table*}[!t]
\renewcommand{\arraystretch}{1}
\setlength\tabcolsep{4.5pt}
\tiny
    \centering
    
    \begin{tabular*}{0.85\textwidth}{@{\extracolsep{\fill}}@{}l l l l c@{}}
    \toprule
    Model ID & Abbreviation & Base Model & Alg. & Size\\

    \midrule
    \multicolumn{4}{l}{\textit{DeepSeek}} \\
    \midrule
     \href{https://huggingface.co/deepseek-ai/DeepSeek-R1-Distill-Qwen-1.5B}{DeepSeek-R1-Distill-Qwen-1.5B} & DS-R1-1.5B & Qwen2.5-Math-1.5B & SFT & 800K\\ 
     \href{https://huggingface.co/deepseek-ai/DeepSeek-R1-Distill-Qwen-7B}{DeepSeek-R1-Distill-Qwen-7B} & DS-R1-7B & Qwen2.5-Math-7B & SFT & 800K\\
     \href{https://huggingface.co/deepseek-ai/DeepSeek-R1-Distill-Llama-8B}{DeepSeek-R1-Distill-Llama-8B}	 & DS-R1-8B & Llama-3.1-8B & SFT & 800K\\
     \href{https://huggingface.co/deepseek-ai/DeepSeek-R1-Distill-Qwen-14B}{DeepSeek-R1-Distill-Qwen-14B} & DS-R1-14B & Qwen2.5-14B & SFT & 800K\\
     \href{https://huggingface.co/deepseek-ai/DeepSeek-R1-Distill-Qwen-32B}{DeepSeek-R1-Distill-Qwen-32B} & DS-R1-32B & Qwen2.5-32B & SFT & 800K\\
     \href{https://huggingface.co/deepseek-ai/DeepSeek-R1-Distill-Llama-70B}{DeepSeek-R1-Distill-Llama-70B} & DS-R1-70B & Llama-3.3-70B-Instruct & SFT & 800K\\
     \href{https://huggingface.co/deepseek-ai/DeepSeek-R1}{DeepSeek-R1} & DS-R1 & DeepSeek-V3-0324 & SFT\&RL & 800K\&--\\

    \midrule
    \multicolumn{4}{l}{\textit{Qwen}} \\
    \midrule
     \href{https://huggingface.co/Qwen/QwQ-32B}{QwQ-32B} & QwQ-32B & Qwen2.5-32B & -- & --\\
     \href{https://huggingface.co/Qwen/Qwen3-235B-A22B}{Qwen3-235B-A22B} & Qwen3-A22B & Qwen3-235B-A22B-Base & SFT\&RL & --\&-- \\
     \href{https://huggingface.co/Qwen/Qwen3-30B-A3B}{Qwen3-30B-A3B} & Qwen3-A3B & Qwen3-30B-A3B-Base & SFT\&RL & --\&-- \\
     \href{https://huggingface.co/Qwen/Qwen3-32B}{Qwen3-32B} & Qwen3-32B & Qwen3-32B-Base & SFT\&RL & --\&-- \\
     \href{https://huggingface.co/Qwen/Qwen3-14B}{Qwen3-14B} & Qwen3-14B & Qwen3-14B-Base & SFT\&RL & --\&-- \\
     \href{https://huggingface.co/Qwen/Qwen3-8B}{Qwen3-8B} & Qwen3-8B & Qwen3-8B-Base & SFT\&RL & --\&-- \\
     \href{https://huggingface.co/Qwen/Qwen3-1.7B}{Qwen3-1.7B} & Qwen3-1.7B & Qwen3-1.7B-Base & SFT\&RL & --\&-- \\

    \midrule
    \multicolumn{4}{l}{\textit{qihoo360}} \\
    \midrule
     \href{https://huggingface.co/qihoo360/Light-R1-7B-DS}{Light-R1-7B-DS} & L-R1-7B-DS & DeepSeek-R1-Distill-Qwen-7B & SFT & 3K\\
     \href{https://huggingface.co/qihoo360/Light-R1-14B-DS}{Light-R1-14B-DS} & L-R1-14B-DS & DeepSeek-R1-Distill-Qwen-14B & SFT\&RL & 3K\&--\\
     \href{https://huggingface.co/qihoo360/Light-R1-32B-DS}{Light-R1-32B-DS} & L-R1-32B-DS & DeepSeek-R1-Distill-Qwen-32B & SFT & 3K\\
     \href{https://huggingface.co/qihoo360/Light-R1-32B}{Light-R1-32B} & L-R1-32B & Qwen2.5-32B-Instruct & SFT\&DPO & 73K\&--\\

    \midrule
    \multicolumn{4}{l}{\textit{simplescaling}} \\
    \midrule 
     \href{https://huggingface.co/simplescaling/s1.1-7B}{s1.1-7B} & s1.1-7B & Qwen2.5-7B-Instruct & SFT & 1K\\
     \href{https://huggingface.co/simplescaling/s1.1-14B}{s1.1-14B} & s1.1-14B & Qwen2.5-14B-Instruct & SFT & 1K\\
     \href{https://huggingface.co/simplescaling/s1.1-32B}{s1.1-32B} & s1.1-32B & Qwen2.5-32B-Instruct & SFT & 1K\\

     
    \midrule
    \multicolumn{4}{l}{\textit{NVIDIA}} \\
    \midrule 
     \href{https://huggingface.co/nvidia/Llama-3.1-Nemotron-Nano-8B-v1}{Llama-3.1-Nemotron-Nano-8B-v1} & Nemotron-8B & Llama-3.1-8B-Instruct & SFT\&RL & --\&--\\
     \href{https://huggingface.co/nvidia/Llama-3.3-Nemotron-Super-49B-v1}{Llama-3.3-Nemotron-Super-49B-v1} & Nemotron-49B & Llama-3.3-70B-Instruct & SFT\&RL & --\&--\\


    \bottomrule
    
    \end{tabular*}
    \caption{The open-source LRMs details evaluated for S1-Bench.}
    \label{tab:model-settings}
\end{table*}

%% file: tables/appendix_count_benchmark.tex
\begin{table*}[h!]
\centering
\footnotesize
\renewcommand{\arraystretch}{1.3}
\setlength\tabcolsep{2.5pt}
    \begin{tabular}{l | cc cc cc cc cc c}
    \toprule
    \multirow{2}{*}{\raisebox{-0.1\height}{\textbf{Paper Abbreviation}}} & 
    \multirow{2}{*}{\raisebox{-0.1\height}{\textbf{MATH}}} &
    \multirow{2}{*}{\raisebox{-0.1\height}{\textbf{GSM8K}}} &
    \multirow{2}{*}{\raisebox{-0.1\height}{\textbf{AIME}}} &
    \multirow{2}{*}{\raisebox{-0.1\height}{\textbf{GPQA}}} & 
    \multirow{2}{*}{\raisebox{-0.1\height}{\textbf{AMC}}} & 
    \multirow{2}{*}{\raisebox{-0.1\height}{\textbf{MMLU}}} &
    \textbf{Olympiad-} & 
    \multirow{2}{*}{\raisebox{-0.1\height}{\textbf{SVAMP}}} &
    \textbf{LiveCode-} &
    \textbf{Common-} \\
    &  &  &  &  &  &  & \textbf{Bench} &  & \textbf{Bench} & \textbf{SenseQA} \\
    \midrule
    Codi~\cite{shen2025codi}  &  & \cmark &  &  &  &  &  & \cmark &  &  \\
    CISC~\cite{taubenfeld2025confidence}  & \cmark & \cmark &  &  &  & \cmark &  &  &  &  \\
    CoT-Valve~\cite{ma2025cot}  &  & \cmark & \cmark &  &  &  &  &  &  &  \\
    Dast~\cite{shen2025dast}  & \cmark &  & \cmark &  &  &  &  &  &  &  \\
    ATM~\cite{chen2024distilling}  &  & \cmark &  &  &  &  &  & \cmark &  & \cmark \\
    DEER~\cite{yang2025dynamic}  & \cmark &  & \cmark & \cmark & \cmark &  &  &  &  &  \\
    DPTS~\cite{ding2025dynamic}  & \cmark & \cmark &  &  &  &  &  &  &  &  \\
    Dynasor~\cite{fu2024efficiently}  &  & \cmark &  &  &  &  &  &  & \cmark &  \\
    ESC~\cite{li2024escape}  & \cmark & \cmark &  &  &  &  &  & \cmark &  & \cmark \\
    Hawkeye~\cite{she2025hawkeye}  & \cmark & \cmark & \cmark &  &  &  &  &  &  &  \\
    token complexity~\cite{lee2025well}  & \cmark & \cmark &  &  &  & \cmark &  &  &  &  \\
    INFTYTHINK~\cite{yan2025inftythink}  & \cmark &  & \cmark & \cmark &  &  &  &  &  &  \\
    KIMI K1.5~\cite{team2501kimi}  & \cmark &  & \cmark &  &  & \cmark &  &  & \cmark &  \\
    L1~\cite{aggarwal2025l1}  & \cmark &  & \cmark & \cmark &  & \cmark & \cmark &  &  &  \\
    LightThinker~\cite{zhang2025lightthinker}  &  & \cmark &  & \cmark &  & \cmark &  &  &  &  \\
    LS-Mixture SFT~\cite{yu2025long}  & \cmark &  & \cmark & \cmark &  &  &  &  &  &  \\
    DSC~\cite{wang2024make}  & \cmark & \cmark &  &  &  &  &  &  &  &  \\
    O1-Pruner~\cite{luo2025o1}  & \cmark & \cmark &  &  &  &  &  &  &  &  \\
    MRT~\cite{qu2025optimizing}  & \cmark &  & \cmark & \cmark &  &  &  &  &  &  \\
    Self-Doubt~\cite{fu2025reasoning}  & \cmark &  & \cmark &  & \cmark &  &  &  &  &  \\
    RASC~\cite{wan2024dynamic}  &  & \cmark &  &  &  &  &  & \cmark &  & \cmark \\
    NoThinking~\cite{ma2025reasoning}  &  &  & \cmark &  & \cmark &  & \cmark &  & \cmark &  \\
    Retro-Search~\cite{lu2025retro}  & \cmark & \cmark & \cmark &  & \cmark &  & \cmark &  &  &  \\
    RSD~\cite{liao2025reward}  & \cmark & \cmark &  & \cmark &  & \cmark & \cmark &  &  &  \\
    ST-BoN~\cite{wang2025sampling}  & \cmark &  &  & \cmark &  &  &  &  &  &  \\
    Elastic Reasoning~\cite{xu2025scalable}  & \cmark &  & \cmark &  & \cmark &  & \cmark &  & \cmark &  \\
    FS-BoN~\cite{munkhbat2502self}  & \cmark & \cmark &  &  &  &  &  &  &  &  \\
    SoT~\cite{aytes2025sketch}  &  & \cmark &  &  &  &  &  & \cmark &  & \cmark \\
    SpecReason~\cite{pan2025specreason}  & \cmark &  & \cmark & \cmark &  &  &  &  &  &  \\
    Speculative Thinking~\cite{yang2025speculative}  & \cmark &  & \cmark & \cmark & \cmark &  &  &  &  &  \\
    SPIRIT~\cite{cui2025stepwise}  &  & \cmark &  &  &  &  &  &  &  &  \\
    ITC Analysis~\cite{wang2025think}  & \cmark &  & \cmark & \cmark &  &  &  &  &  &  \\
    Think when needed~\cite{yang2025think}  &  &  & \cmark &  &  &  &  &  &  &  \\
    THINKPRUNE~\cite{hou2025thinkprune}  & \cmark &  & \cmark &  & \cmark &  & \cmark &  &  &  \\
    TALE~\cite{han2024token}  &  & \cmark &  &  &  &  &  &  &  &  \\
    TokenSkip~\cite{xia2025tokenskip}  & \cmark & \cmark &  &  &  &  &  &  &  &  \\
    TOPS~\cite{yang2025towards}  & \cmark & \cmark & \cmark &  &  &  &  &  &  &  \\
    efficient reasoning~\cite{arora2025training}  & \cmark & \cmark & \cmark &  &  &  &  &  &  &  \\
    TWT~\cite{xu2025twt}  &  &  &  &  &  &  &  &  &  & \cmark \\
    Z1~\cite{yu2025z1}  & \cmark &  &  & \cmark &  &  &  &  & \cmark &  \\
    \midrule
    Count & 28 & 24 & 20 & 11 & 8 & 6 & 6 & 5 & 5 & 5 \\
    
\bottomrule
\end{tabular}
\caption{A total of 40 studies on LRM efficiency before May 2025 were included. Benchmarks that appeared more than four times are listed.}
\label{app:count_benchmark}
\end{table*}

%% file: tables/st_bench_sub_cat_intro.tex
\begin{table*}[!h]

\renewcommand{\arraystretch}{1.2}
\setlength\tabcolsep{5pt}
\small
    \centering
    \begin{tabular}{c | c | l }
    \toprule
    \textbf{cate.} & \textbf{subcatrgories} & \textbf{Explanation and cases} \\
    \midrule
    \multirow{23}{*}{\rotatebox[origin=c]{90}{\textbf{reasoning question}}}
        & \multirow{3}{*}{numerical reasoning} 
            &   Questions that require performing basic mathematical operations or solving simple algebraic  \\
            & & equations to arrive at a numerical answer. \\
            & & \textbf{Case:} What's two plus three?  \\
            \cmidrule(lr){2-3}
        & \multirow{4}{*}{code reasoning} 
            &   Questions that require tracing through and executing simple code snippets to determine their  \\
            & & output or behavior when run in a specific programming environment. \\
            & & \textbf{Case:} What is the output of the following code when run in Python 3 environment: word =  \\
            & & "hello"\textbackslash nprint(len(word)) \\
            \cmidrule(lr){2-3}
        & \multirow{3}{*}{set reasoning}
            &   Questions that require applying simple syllogistic reasoning to determine whether elements  \\
            & & belong to sets based on clearly stated relationships.  \\
            & & \textbf{Case:} All squares are quadrilaterals. A shape is a square, is it a quadrilateral?  \\
            \cmidrule(lr){2-3}
        & \multirow{3}{*}{temporal reasoning}
            &   Questions that require calculating time durations, ages, or future dates by applying simple   \\
            & & arithmetic operations to temporal information.   \\
            & & \textbf{Case:} How many minutes equal 120 seconds?  \\
            \cmidrule(lr){2-3}
        & \multirow{3}{*}{spatial reasoning}
            &   Questions that require determining relative positions, directions, or orientations of objects  \\
            & & in space based on simple spatial relationships.   \\
            & & \textbf{Case:} If a bird is flying above a tree, where is the tree in relation to the bird?  \\
            \cmidrule(lr){2-3}
        & \multirow{4}{*}{causal reasoning}
            &   Questions that require determining outcomes by applying simple cause-and-effect   \\
            & & relationships based on given conditional statements.   \\
            & & \textbf{Case:} If ferromagnetic material is placed in a magnetic field, it will become magnetized. An   \\
            & & iron nail was placed next to a strong magnet for some time. Has the nail been magnetized? \\
            \cmidrule(lr){2-3}
        & \multirow{3}{*}{natural law reasoning}
            &   Questions that require applying basic knowledge of physical laws and natural phenomena to   \\
            & & predict simple observable outcomes in everyday scenarios.   \\
            & & \textbf{Case:} Which is faster, an airplane or the propagation of light?  \\

    \midrule
    \multirow{20}{*}{\rotatebox[origin=c]{90}{\textbf{knowledge question}}}
    & \multirow{3}{*}{geometry facts}
            &   Questions that require recalling simple and fundamental geometric properties about shapes,    \\
            & & angles, and basic geometric figures.    \\
            & & \textbf{Case:} How many angles does a trapezoid have?  \\
            \cmidrule(lr){2-3}
    & \multirow{3}{*}{geographic facts}
            &   Questions that require recalling simple factual information about locations, landmarks,     \\
            & & political divisions, celestial bodies, and other basic geographic knowledge.   \\
            & & \textbf{Case:} Which is the largest continent on Earth?   \\
            \cmidrule(lr){2-3}
    & \multirow{2}{*}{historical facts}
            &   Questions that require recalling basic facts about historical events.    \\
            & & \textbf{Case:} Which country first invented paper?  \\
            \cmidrule(lr){2-3}
    & \multirow{3}{*}{biographical facts}
            &   Questions that require recalling basic facts about the identities, achievements, and    \\
            & & characteristics of historical figures.   \\
            & & \textbf{Case:} Who proposed the theory of universal gravitation?  \\
            \cmidrule(lr){2-3}
    & \multirow{3}{*}{measurement units}
            &   Questions that require recalling simple conversion relationships between     \\
            & & standard units of measurement.   \\
            & & \textbf{Case:} How many centimeters equal 1 meter?  \\
            \cmidrule(lr){2-3}
    & \multirow{3}{*}{scientific notation}
            &   Questions that require recalling basic scientific symbols, formulas, and standard units    \\
            & & used in scientific communication.    \\
            & & \textbf{Case:} What is the chemical symbol for oxygen?  \\
            \cmidrule(lr){2-3}
    & \multirow{3}{*}{creative authorship}
            &   Questions that require recalling the creators or originators of notable artistic, literary,     \\
            & & musical, and cultural works.   \\
            & & \textbf{Case:} Who is the author of Hamlet?  \\

   \bottomrule
   
    \end{tabular}
    \caption{The subcategory descriptions and cases of reasoning questions and knowledge questions.}
    \label{tabel:sub_cat_exp_1}
\end{table*}


%% file: tables/st_bench_sub_cat_intro_2.tex
\begin{table*}[!h]

\renewcommand{\arraystretch}{1.2}
\setlength\tabcolsep{5pt}
\small
    \centering
    \begin{tabular}{c | c | l }
    \toprule
    \textbf{cate.} & \textbf{subcatrgories} & \textbf{Explanation and cases} \\
    \midrule
    \multirow{18}{*}{\rotatebox[origin=c]{90}{\textbf{instruction following}}}
    & \multirow{3}{*}{repetition constraints}
        &   Questions that require outputting specified characters, words, or phrases a specific number  \\
        & & of times according to simple formatting instructions.   \\
        & & \textbf{Case:} Output the number "7" four times, without using separators.  \\
        \cmidrule(lr){2-3}
    & \multirow{3}{*}{length constraints}
        &   Questions that require generating outputs of a specific length or with a specific number \\
        & & of components based on simple counting constraints.   \\
        & & \textbf{Case:} Output a four-digit number.  \\
        \cmidrule(lr){2-3}
    & \multirow{3}{*}{character constraints}
        &   Questions that require generating words or numbers that conform to simple specified   \\
        & & character patterns or formatting rules.  \\
        & & \textbf{Case:} Output a number that begins with 8.  \\
        \cmidrule(lr){2-3}
    & \multirow{3}{*}{counting constraints}
        &   Questions that require counting specific characters or elements within   \\
        & & a given text or sequence.  \\
        & & \textbf{Case:} Output the number of letter "y" in the word "yes".  \\
        \cmidrule(lr){2-3}
    & \multirow{3}{*}{transformation constraints}
        &   Questions that require modifying text or numbers according to simple formatting or character   \\
        & & substitution rules to produce a transformed output.  \\
        & & \textbf{Case:} Output the word "good" with all letters capitalized directly.  \\
        \cmidrule(lr){2-3}
    & \multirow{3}{*}{sentence constraints}
        &   Questions that require generating sentences that conform to simple specified content or   \\
        & & structural requirements.  \\
        & & \textbf{Case:} Give a sentence that contains the phrase "have lunch" directly.  \\

    \midrule
    \multirow{24}{*}{\rotatebox[origin=c]{90}{\textbf{analysis question}}}
        & \multirow{3}{*}{sentiment classification}
        &   Questions that require determining whether simple statements express positive or negative   \\
        & & emotions based on the tone and word choice.  \\
        & & \textbf{Case:} Does the sentence "I hate rainy days." express a positive or negative emotion?  \\
        \cmidrule(lr){2-3}
    & \multirow{3}{*}{named entity recognition}
        &   Questions that require identifying the correct category of named entities (such as people,   \\
        & & places, organizations, or time expressions) within simple sentences.  \\
        & & \textbf{Case:} In the sentence "Napoleon died in 1821", is "1821" a time or a place name?  \\
        \cmidrule(lr){2-3}
    & \multirow{3}{*}{language classification}
        &   Questions that require identifying the language of origin for simple words or phrases   \\
        & & based on their characteristic writing systems or common vocabulary.  \\
        & & \textbf{Case:} Is the word "hello" English or Japanese?  \\
        \cmidrule(lr){2-3}
    & \multirow{3}{*}{topic classification}
        &   Questions that require identifying the primary subject matter or thematic category of  \\
        & & simple sentences based on their content and context clues.   \\
        & & \textbf{Case:} Is the topic of the sentence "The stock market rose 2\% today" finance or technology?   \\
        \cmidrule(lr){2-3}
    & \multirow{3}{*}{intent recognition}
        &   Questions that require determining the communicative purpose behind simple utterances   \\
        & & or statements based on their phrasing and context.  \\
        & & \textbf{Case:} Is the intention of the sentence "I'm sorry I'm late." to apologize or to blame?  \\
        \cmidrule(lr){2-3}
    & \multirow{3}{*}{syntax classification}
        &   Questions that require identifying the correct grammatical structure or sentence type of  \\
        & & simple expressions based on their form, punctuation, and communicative function.   \\
        & & \textbf{Case:} Is "Close the door!" an imperative sentence or an interrogative sentence?   \\
        \cmidrule(lr){2-3}
    & \multirow{3}{*}{grammar classification}
        &   Questions that require identifying simple grammatical properties (like tense, voice,   \\
        & & or polarity) of sentences based on their structure and verb forms.  \\
        & & \textbf{Case:} Is "The apple was eaten." in active voice or passive voice?  \\
        \cmidrule(lr){2-3}
    & \multirow{3}{*}{coreference resolution}
        &   Questions that require identifying which entity a pronoun or reference term refers to in   \\
        & & simple sentences by tracking relationships between words in the text.  \\
        & & \textbf{Case:} In "My computer is broken, and I need to fix it." What does "it" refer to?  \\

    \bottomrule
    \end{tabular}
    \caption{The subcategory descriptions and cases of instruction following questions and analysis questions.}
    \label{tabel:sub_cat_exp_2}
\end{table*}

%% file: prompt/S1_bench_workflow.tex
\begin{table*}[!t]
\small
\centering
\begin{tabular}{@{}p{1.0\linewidth}@{}}
\toprule
\textbf{Prompt for construction workflow for S1-Bench}\\
\midrule
\textit{\textbf{Data Generation Prompt}}\\
\midrule


Generate 50 pairs of questions and answers in both Chinese and English based on the category's name, definition, and specific simplicity criteria. The following conditions must be satisfied:\\
1. Questions must be naturally and clearly expressed, unambiguous, and free of intentional traps.\\
2. Answers must be unique or easily falsifiable, with no possibility of multiple correct answers.\\
3. Make the questions as diverse as possible.\\
\vspace{-1pt}

\# Category Name and Definition:\\
\{name\_and\_definition\}\\
\vspace{-1pt}

\# Specific Simplicity Criteria:\\
\{criteria\}\\
\vspace{-1pt}

\# Cases:\\
\#\# English question:\\
\{question\_en\}\\
\#\# English Answer:\\
\{answer\_en\}\\
\vspace{-1pt}

\#\# Chinese question:\\
\{question\_zh\}\\
\#\# Chinese Answer:\\
\{answer\_zh\}\\
\vspace{-1pt}

Please generate 50 pairs of Chinese and English questions and answers in the following format:

[question]English-question[answer]English-answer[question]Chinese-question[answer]Chinese-answer...\\
\vspace{-1pt}

Start generating:\\

\midrule
\textit{\textbf{Quality Discrimination Prompt}}\\
\midrule

Given a question, its answer, and its category, please analyze from the following perspectives as comprehensively as possible:\\
1. Whether the question belongs to the specified category and meet the Specific Simplicity Criteria.\\
2. Whether the question is easy, clear, unambiguous, and has an absolutely unique answer.\\
3. Whether the answer is absolutely correct; if not, what the correct answer should be.\\
4. Whether the question is similar to other given questions, and if similar, whether more diverse questions can be generated.\\
\vspace{-1pt}

\# Category Name and Definition:\\
\{name\_and\_definition\}\\
\vspace{-1pt}

\# Specific Simplicity Criteria:\\
\{criteria\}\\
\vspace{-1pt}

\# Question and Answer:\\
\{question\_with\_answer\}\\
\vspace{-1pt}

\# Other Questions:\\
\{questions\_list\}\\
\vspace{-1pt}

Begin your analysis, aiming to be as detailed and comprehensive as possible:\\

\midrule
\textit{\textbf{Difficulty Reduction Prompt}}\\
\midrule

Given a question and answer that are too complex for the model to answer correctly, you need to further reduce their difficulty while trying to:\\
- Ensure the question aligns with the Category Name and Definition. \\
- Ensure the question meets the Specific Simplicity Criteria. \\
\vspace{-1pt}

\# Category Name and Definition:\\
\{name\_and\_definition\}\\
\vspace{-1pt}

\# Specific Simplicity Criteria:\\
\{criteria\}\\
\vspace{-1pt}

\# Question and Answer:\\
\{question\_with\_answer\}\\
\vspace{-1pt}

The new question and answer:\\

\bottomrule
\end{tabular}
    \caption{“Category Name and Definition” refers to the subcategory name and its definition, while Specific Simplicity Criteria refers to the simplicity requirements specific to the main category.}
    \label{table:prompt_s1_workflow}
\end{table*}

%% file: prompt/eval_4o.tex
\begin{table*}[!t]
\small
\renewcommand{\arraystretch}{1.0}
\centering
\begin{tabular}{@{}p{1.0\linewidth}@{}}
\toprule
\textbf{Prompt for Correctness Evaluation}\\
\midrule
\textit{\textbf{Evaluation on S1-Bench}}\\
\midrule

**Question:** \\
\{question\}\\
\vspace{-1pt}

**Model Answer:**\\
\{model\_answer\}\\
\vspace{-1pt}

**Ground Truth:** \\
\{ground\_truth\}\\
\vspace{-1pt}

Your task is to evaluate whether the model's answer is correct. \\
An answer is considered correct as long as it contains the ground truth (regardless of how complex or detailed the description is).\\
If there are parenthetical notes after the ground truth, then there may be multiple correct answers. In this case, the given answer is just one example, and any answer that meets the requirements specified in the notes can be considered correct.\\
Additionally, some reasonably uncertain supplementary information is also considered appropriate, including more details, possibilities, and expanded discussion. You should focus more on whether the reply contains the correct answer.\\
\vspace{-1pt}

You need to output a standard JSON, providing your explanation of the evaluation in the "explain" field, and giving the evaluation result in the "result" field, where 1 means the answer is correct and 0 means it is incorrect.\\
Your action should follow the given format: {{``explain": ``", ``result": 0/1}}\\

\bottomrule
\end{tabular}
    \caption{Prompt for Correctness Evaluation.}
    \label{prompt:eval_4o}
\end{table*}

%% file: figure/exp_hot.tex
\begin{figure*}[!h]  
\centering  
\includegraphics[width=15cm]{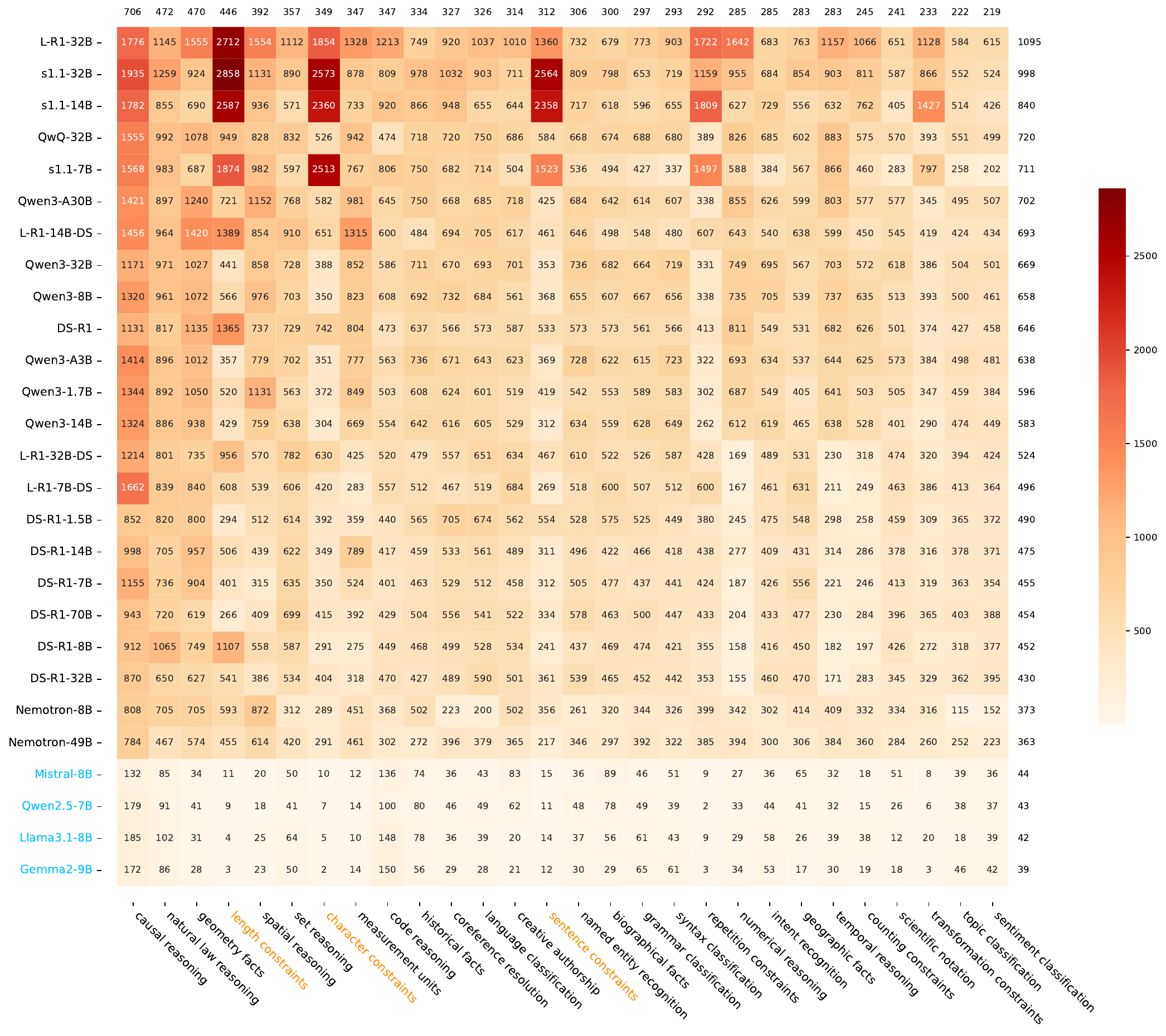}
\caption{Average response token counts on the 28 subcategories, which is the average result of five generations under top-p sampling.}
\label{fig:hot}
\end{figure*}

%% file: tables/exp_error_format_number_temp.tex
\begin{table*}[!t]
\renewcommand{\arraystretch}{1.0}
\small
    \centering
    
    \begin{tabular*}{0.8\textwidth}{@{\extracolsep{\fill}}@{}l | cc | cccc cccc | cc @{}}

     \toprule
   \multirow{2}{*}{\raisebox{-0.2\height}{Model}} & \multicolumn{2}{c|}{\raisebox{-0.2\height}{\textbf{Standard}}} & \multicolumn{8}{c|}{\raisebox{-0.2\height}{\textbf{Readable but Malformed}}} &  \multicolumn{2}{c}{\raisebox{-0.2\height}{\textbf{Unreadable}}} \\
    \cmidrule(lr){2-13}  
    & 100 & 101 & 200 & 201 & 202 & 203 & 204 & 205 & 206 & 207 & 300 & 301 \\
    
    \midrule
Qwen3-A22B & 100.00 & \textcolor{whitegray}{0.00} & \textcolor{whitegray}{0.00} & \textcolor{whitegray}{0.00} & \textcolor{whitegray}{0.00} & \textcolor{whitegray}{0.00} & \textcolor{whitegray}{0.00} & \textcolor{whitegray}{0.00} & \textcolor{whitegray}{0.00} & \textcolor{whitegray}{0.00} & \textcolor{whitegray}{0.00} & \textcolor{whitegray}{0.00}  \\
Qwen3-A3B & 100.00 & \textcolor{whitegray}{0.00} & \textcolor{whitegray}{0.00} & \textcolor{whitegray}{0.00} & \textcolor{whitegray}{0.00} & \textcolor{whitegray}{0.00} & \textcolor{whitegray}{0.00} & \textcolor{whitegray}{0.00} & \textcolor{whitegray}{0.00} & \textcolor{whitegray}{0.00} & \textcolor{whitegray}{0.00} & \textcolor{whitegray}{0.00}  \\
QwQ-32B & 100.00 & \textcolor{whitegray}{0.00} & \textcolor{whitegray}{0.00} & \textcolor{whitegray}{0.00} & \textcolor{whitegray}{0.00} & \textcolor{whitegray}{0.00} & \textcolor{whitegray}{0.00} & \textcolor{whitegray}{0.00} & \textcolor{whitegray}{0.00} & \textcolor{whitegray}{0.00} & \textcolor{whitegray}{0.00} & \textcolor{whitegray}{0.00}  \\
Qwen3-32B & 99.91 & \textcolor{whitegray}{0.00} & \textcolor{whitegray}{0.00} & \textcolor{whitegray}{0.00} & \textcolor{whitegray}{0.00} & \textcolor{whitegray}{0.00} & \textcolor{whitegray}{0.00} & \textcolor{whitegray}{0.00} & \textcolor{whitegray}{0.00} & \textcolor{whitegray}{0.00} & \textcolor{whitegray}{0.00} & 0.09  \\
Qwen3-14B & 99.95 & \textcolor{whitegray}{0.00} & \textcolor{whitegray}{0.00} & \textcolor{whitegray}{0.00} & \textcolor{whitegray}{0.00} & \textcolor{whitegray}{0.00} & \textcolor{whitegray}{0.00} & \textcolor{whitegray}{0.00} & \textcolor{whitegray}{0.00} & \textcolor{whitegray}{0.00} & \textcolor{whitegray}{0.00} & 0.05  \\
Qwen3-8B & 99.95 & \textcolor{whitegray}{0.00} & \textcolor{whitegray}{0.00} & \textcolor{whitegray}{0.00} & \textcolor{whitegray}{0.00} & \textcolor{whitegray}{0.00} & \textcolor{whitegray}{0.00} & \textcolor{whitegray}{0.00} & \textcolor{whitegray}{0.00} & \textcolor{whitegray}{0.00} & \textcolor{whitegray}{0.00} & 0.05  \\
Qwen3-1.7B & 99.81 & \textcolor{whitegray}{0.00} & \textcolor{whitegray}{0.00} & \textcolor{whitegray}{0.00} & \textcolor{whitegray}{0.00} & \textcolor{whitegray}{0.00} & \textcolor{whitegray}{0.00} & \textcolor{whitegray}{0.00} & \textcolor{whitegray}{0.00} & \textcolor{whitegray}{0.00} & \textcolor{whitegray}{0.00} & 0.19  \\


DS-R1 & 100.00 & \textcolor{whitegray}{0.00} & \textcolor{whitegray}{0.00} & \textcolor{whitegray}{0.00} & \textcolor{whitegray}{0.00} & \textcolor{whitegray}{0.00} & \textcolor{whitegray}{0.00} & \textcolor{whitegray}{0.00} & \textcolor{whitegray}{0.00} & \textcolor{whitegray}{0.00} & \textcolor{whitegray}{0.00} & \textcolor{whitegray}{0.00}  \\
DS-R1-70B & 99.91 & \textcolor{whitegray}{0.00} & 0.09 & \textcolor{whitegray}{0.00} & \textcolor{whitegray}{0.00} & \textcolor{whitegray}{0.00} & \textcolor{whitegray}{0.00} & \textcolor{whitegray}{0.00} & \textcolor{whitegray}{0.00} & \textcolor{whitegray}{0.00} & \textcolor{whitegray}{0.00} & \textcolor{whitegray}{0.00}  \\
DS-R1-32B & 100.00 & \textcolor{whitegray}{0.00} & \textcolor{whitegray}{0.00} & \textcolor{whitegray}{0.00} & \textcolor{whitegray}{0.00} & \textcolor{whitegray}{0.00} & \textcolor{whitegray}{0.00} & \textcolor{whitegray}{0.00} & \textcolor{whitegray}{0.00} & \textcolor{whitegray}{0.00} & \textcolor{whitegray}{0.00} & \textcolor{whitegray}{0.00}  \\
DS-R1-14B & 100.00 & \textcolor{whitegray}{0.00} & \textcolor{whitegray}{0.00} & \textcolor{whitegray}{0.00} & \textcolor{whitegray}{0.00} & \textcolor{whitegray}{0.00} & \textcolor{whitegray}{0.00} & \textcolor{whitegray}{0.00} & \textcolor{whitegray}{0.00} & \textcolor{whitegray}{0.00} & \textcolor{whitegray}{0.00} & \textcolor{whitegray}{0.00}  \\
DS-R1-8B & 99.53 & \textcolor{whitegray}{0.00} & \textcolor{whitegray}{0.00} & \textcolor{whitegray}{0.00} & \textcolor{whitegray}{0.00} & \textcolor{whitegray}{0.00} & \textcolor{whitegray}{0.00} & \textcolor{whitegray}{0.00} & \textcolor{whitegray}{0.00} & 0.24 & \textcolor{whitegray}{0.00} & 0.24  \\
DS-R1-7B & 99.24 & \textcolor{whitegray}{0.00} & \textcolor{whitegray}{0.00} & \textcolor{whitegray}{0.00} & \textcolor{whitegray}{0.00} & \textcolor{whitegray}{0.00} & \textcolor{whitegray}{0.00} & \textcolor{whitegray}{0.00} & \textcolor{whitegray}{0.00} & \textcolor{whitegray}{0.00} & \textcolor{whitegray}{0.00} & 0.76  \\
DS-R1-1.5B & 97.58 & \textcolor{whitegray}{0.00} & \textcolor{whitegray}{0.00} & \textcolor{whitegray}{0.00} & \textcolor{whitegray}{0.00} & \textcolor{whitegray}{0.00} & \textcolor{whitegray}{0.00} & \textcolor{whitegray}{0.00} & \textcolor{whitegray}{0.00} & \textcolor{whitegray}{0.00} & \textcolor{whitegray}{0.00} & 2.42  \\


Nemotron-49B & 66.07 & 33.93 & \textcolor{whitegray}{0.00} & \textcolor{whitegray}{0.00} & \textcolor{whitegray}{0.00} & \textcolor{whitegray}{0.00} & \textcolor{whitegray}{0.00} & \textcolor{whitegray}{0.00} & \textcolor{whitegray}{0.00} & \textcolor{whitegray}{0.00} & \textcolor{whitegray}{0.00} & \textcolor{whitegray}{0.00}  \\
Nemotron-8B & 58.06 & 26.26 & \textcolor{whitegray}{0.00} & \textcolor{whitegray}{0.00} & \textcolor{whitegray}{0.00} & 0.09 & \textcolor{whitegray}{0.00} & \textcolor{whitegray}{0.00} & \textcolor{whitegray}{0.00} & 15.02 & \textcolor{whitegray}{0.00} & 0.57  \\

L-R1-32B & 95.07 & \textcolor{whitegray}{0.00} & \textcolor{whitegray}{0.00} & \textcolor{whitegray}{0.00} & \textcolor{whitegray}{0.00} & \textcolor{whitegray}{0.00} & 0.81 & \textcolor{whitegray}{0.00} & \textcolor{whitegray}{0.00} & 3.03 & \textcolor{whitegray}{0.00} & 1.09  \\
L-R1-32B-DS & 99.81 & \textcolor{whitegray}{0.00} & \textcolor{whitegray}{0.00} & \textcolor{whitegray}{0.00} & \textcolor{whitegray}{0.00} & \textcolor{whitegray}{0.00} & \textcolor{whitegray}{0.00} & \textcolor{whitegray}{0.00} & \textcolor{whitegray}{0.00} & \textcolor{whitegray}{0.00} & \textcolor{whitegray}{0.00} & 0.19  \\
L-R1-14B-DS & 99.19 & \textcolor{whitegray}{0.00} & \textcolor{whitegray}{0.00} & \textcolor{whitegray}{0.00} & \textcolor{whitegray}{0.00} & \textcolor{whitegray}{0.00} & \textcolor{whitegray}{0.00} & \textcolor{whitegray}{0.00} & \textcolor{whitegray}{0.00} & \textcolor{whitegray}{0.00} & \textcolor{whitegray}{0.00} & 0.81  \\
L-R1-7B-DS & 99.67 & \textcolor{whitegray}{0.00} & 0.05 & 0.05 & \textcolor{whitegray}{0.00} & \textcolor{whitegray}{0.00} & \textcolor{whitegray}{0.00} & \textcolor{whitegray}{0.00} & \textcolor{whitegray}{0.00} & \textcolor{whitegray}{0.00} & \textcolor{whitegray}{0.00} & 0.24  \\

s1.1-32B & 99.53 & \textcolor{whitegray}{0.00} & \textcolor{whitegray}{0.00} & 0.05 & \textcolor{whitegray}{0.00} & \textcolor{whitegray}{0.00} & \textcolor{whitegray}{0.00} & \textcolor{whitegray}{0.00} & \textcolor{whitegray}{0.00} & \textcolor{whitegray}{0.00} & \textcolor{whitegray}{0.00} & 0.43  \\
s1.1-14B & 97.39 & \textcolor{whitegray}{0.00} & \textcolor{whitegray}{0.00} & 0.14 & \textcolor{whitegray}{0.00} & \textcolor{whitegray}{0.00} & 0.24 & \textcolor{whitegray}{0.00} & \textcolor{whitegray}{0.00} & \textcolor{whitegray}{0.00} & \textcolor{whitegray}{0.00} & 2.23  \\
s1.1-7B & 88.96 & \textcolor{whitegray}{0.00} & \textcolor{whitegray}{0.00} & 7.96 & 0.09 & \textcolor{whitegray}{0.00} & \textcolor{whitegray}{0.00} & \textcolor{whitegray}{0.00} & \textcolor{whitegray}{0.00} & 0.09 & \textcolor{whitegray}{0.00} & 2.89  \\


    \bottomrule
    
    \end{tabular*}
    \caption{Format type rates under top-p sampling.}
    \label{table:error_counts_rates_high_t}
\end{table*}

\begin{table*}[!t]
\renewcommand{\arraystretch}{1.0}
\small
    \centering
    
    \begin{tabular*}{0.8\textwidth}{@{\extracolsep{\fill}}@{}l | cc | cccc cccc | cc @{}}

    \toprule
   \multirow{2}{*}{\raisebox{-0.2\height}{Model}} & \multicolumn{2}{c|}{\raisebox{-0.2\height}{\textbf{Standard}}} & \multicolumn{8}{c|}{\raisebox{-0.2\height}{\textbf{Readable but Malformed}}} &  \multicolumn{2}{c}{\raisebox{-0.2\height}{\textbf{Unreadable}}} \\
    \cmidrule(lr){2-13}  
    & 100 & 101 & 200 & 201 & 202 & 203 & 204 & 205 & 206 & 207 & 300 & 301 \\
    \midrule
    
Qwen3-A22B & 100.00 & \textcolor{whitegray}{0.00} & \textcolor{whitegray}{0.00} & \textcolor{whitegray}{0.00} & \textcolor{whitegray}{0.00} & \textcolor{whitegray}{0.00} & \textcolor{whitegray}{0.00} & \textcolor{whitegray}{0.00} & \textcolor{whitegray}{0.00} & \textcolor{whitegray}{0.00} & \textcolor{whitegray}{0.00} & \textcolor{whitegray}{0.00}  \\
Qwen3-A3B & 100.00 & \textcolor{whitegray}{0.00} & \textcolor{whitegray}{0.00} & \textcolor{whitegray}{0.00} & \textcolor{whitegray}{0.00} & \textcolor{whitegray}{0.00} & \textcolor{whitegray}{0.00} & \textcolor{whitegray}{0.00} & \textcolor{whitegray}{0.00} & \textcolor{whitegray}{0.00} & \textcolor{whitegray}{0.00} & \textcolor{whitegray}{0.00}  \\
QwQ-32B & 100.00 & \textcolor{whitegray}{0.00} & \textcolor{whitegray}{0.00} & \textcolor{whitegray}{0.00} & \textcolor{whitegray}{0.00} & \textcolor{whitegray}{0.00} & \textcolor{whitegray}{0.00} & \textcolor{whitegray}{0.00} & \textcolor{whitegray}{0.00} & \textcolor{whitegray}{0.00} & \textcolor{whitegray}{0.00} & \textcolor{whitegray}{0.00}  \\
Qwen3-32B & 99.76 & \textcolor{whitegray}{0.00} & \textcolor{whitegray}{0.00} & \textcolor{whitegray}{0.00} & \textcolor{whitegray}{0.00} & \textcolor{whitegray}{0.00} & \textcolor{whitegray}{0.00} & \textcolor{whitegray}{0.00} & \textcolor{whitegray}{0.00} & \textcolor{whitegray}{0.00} & \textcolor{whitegray}{0.00} & 0.24  \\
Qwen3-14B & 100.00 & \textcolor{whitegray}{0.00} & \textcolor{whitegray}{0.00} & \textcolor{whitegray}{0.00} & \textcolor{whitegray}{0.00} & \textcolor{whitegray}{0.00} & \textcolor{whitegray}{0.00} & \textcolor{whitegray}{0.00} & \textcolor{whitegray}{0.00} & \textcolor{whitegray}{0.00} & \textcolor{whitegray}{0.00} & \textcolor{whitegray}{0.00}  \\
Qwen3-8B & 100.00 & \textcolor{whitegray}{0.00} & \textcolor{whitegray}{0.00} & \textcolor{whitegray}{0.00} & \textcolor{whitegray}{0.00} & \textcolor{whitegray}{0.00} & \textcolor{whitegray}{0.00} & \textcolor{whitegray}{0.00} & \textcolor{whitegray}{0.00} & \textcolor{whitegray}{0.00} & \textcolor{whitegray}{0.00} & \textcolor{whitegray}{0.00}  \\
Qwen3-1.7B & 99.76 & \textcolor{whitegray}{0.00} & \textcolor{whitegray}{0.00} & \textcolor{whitegray}{0.00} & \textcolor{whitegray}{0.00} & \textcolor{whitegray}{0.00} & \textcolor{whitegray}{0.00} & \textcolor{whitegray}{0.00} & \textcolor{whitegray}{0.00} & \textcolor{whitegray}{0.00} & \textcolor{whitegray}{0.00} & 0.24  \\


DS-R1 & 100.00 & \textcolor{whitegray}{0.00} & \textcolor{whitegray}{0.00} & \textcolor{whitegray}{0.00} & \textcolor{whitegray}{0.00} & \textcolor{whitegray}{0.00} & \textcolor{whitegray}{0.00} & \textcolor{whitegray}{0.00} & \textcolor{whitegray}{0.00} & \textcolor{whitegray}{0.00} & \textcolor{whitegray}{0.00} & \textcolor{whitegray}{0.00}  \\
DS-R1-70B & 100.00 & \textcolor{whitegray}{0.00} & \textcolor{whitegray}{0.00} & \textcolor{whitegray}{0.00} & \textcolor{whitegray}{0.00} & \textcolor{whitegray}{0.00} & \textcolor{whitegray}{0.00} & \textcolor{whitegray}{0.00} & \textcolor{whitegray}{0.00} & \textcolor{whitegray}{0.00} & \textcolor{whitegray}{0.00} & \textcolor{whitegray}{0.00}  \\
DS-R1-32B & 100.00 & \textcolor{whitegray}{0.00} & \textcolor{whitegray}{0.00} & \textcolor{whitegray}{0.00} & \textcolor{whitegray}{0.00} & \textcolor{whitegray}{0.00} & \textcolor{whitegray}{0.00} & \textcolor{whitegray}{0.00} & \textcolor{whitegray}{0.00} & \textcolor{whitegray}{0.00} & \textcolor{whitegray}{0.00} & \textcolor{whitegray}{0.00}  \\
DS-R1-14B & 99.76 & \textcolor{whitegray}{0.00} & \textcolor{whitegray}{0.00} & \textcolor{whitegray}{0.00} & \textcolor{whitegray}{0.00} & \textcolor{whitegray}{0.00} & \textcolor{whitegray}{0.00} & \textcolor{whitegray}{0.00} & \textcolor{whitegray}{0.00} & \textcolor{whitegray}{0.00} & \textcolor{whitegray}{0.00} & 0.24  \\
DS-R1-8B & 99.53 & \textcolor{whitegray}{0.00} & \textcolor{whitegray}{0.00} & \textcolor{whitegray}{0.00} & \textcolor{whitegray}{0.00} & \textcolor{whitegray}{0.00} & \textcolor{whitegray}{0.00} & \textcolor{whitegray}{0.00} & \textcolor{whitegray}{0.00} & 0.24 & \textcolor{whitegray}{0.00} & 0.24  \\
DS-R1-7B & 97.87 & \textcolor{whitegray}{0.00} & \textcolor{whitegray}{0.00} & \textcolor{whitegray}{0.00} & \textcolor{whitegray}{0.00} & \textcolor{whitegray}{0.00} & \textcolor{whitegray}{0.00} & \textcolor{whitegray}{0.00} & \textcolor{whitegray}{0.00} & \textcolor{whitegray}{0.00} & \textcolor{whitegray}{0.00} & 2.13  \\
DS-R1-1.5B & 91.94 & \textcolor{whitegray}{0.00} & \textcolor{whitegray}{0.00} & \textcolor{whitegray}{0.00} & \textcolor{whitegray}{0.00} & \textcolor{whitegray}{0.00} & \textcolor{whitegray}{0.00} & \textcolor{whitegray}{0.00} & \textcolor{whitegray}{0.00} & \textcolor{whitegray}{0.00} & \textcolor{whitegray}{0.00} & 8.06  \\


Nemotron-49B & 60.90 & 39.10 & \textcolor{whitegray}{0.00} & \textcolor{whitegray}{0.00} & \textcolor{whitegray}{0.00} & \textcolor{whitegray}{0.00} & \textcolor{whitegray}{0.00} & \textcolor{whitegray}{0.00} & \textcolor{whitegray}{0.00} & \textcolor{whitegray}{0.00} & \textcolor{whitegray}{0.00} & \textcolor{whitegray}{0.00}  \\
Nemotron-8B & 55.21 & 26.78 & \textcolor{whitegray}{0.00} & \textcolor{whitegray}{0.00} & \textcolor{whitegray}{0.00} & \textcolor{whitegray}{0.00} & \textcolor{whitegray}{0.00} & \textcolor{whitegray}{0.00} & \textcolor{whitegray}{0.00} & 16.35 & \textcolor{whitegray}{0.00} & 1.66  \\

L-R1-32B & 85.55 & 0.24 & \textcolor{whitegray}{0.00} & 0.24 & 0.71 & 0.24 & 0.95 & \textcolor{whitegray}{0.00} & \textcolor{whitegray}{0.00} & 6.64 & 2.61 & 2.84  \\
L-R1-32B-DS & 99.29 & \textcolor{whitegray}{0.00} & \textcolor{whitegray}{0.00} & \textcolor{whitegray}{0.00} & \textcolor{whitegray}{0.00} & \textcolor{whitegray}{0.00} & \textcolor{whitegray}{0.00} & \textcolor{whitegray}{0.00} & \textcolor{whitegray}{0.00} & \textcolor{whitegray}{0.00} & \textcolor{whitegray}{0.00} & 0.71  \\
L-R1-14B-DS & 98.82 & \textcolor{whitegray}{0.00} & \textcolor{whitegray}{0.00} & \textcolor{whitegray}{0.00} & \textcolor{whitegray}{0.00} & \textcolor{whitegray}{0.00} & \textcolor{whitegray}{0.00} & \textcolor{whitegray}{0.00} & \textcolor{whitegray}{0.00} & \textcolor{whitegray}{0.00} & \textcolor{whitegray}{0.00} & 1.18  \\
L-R1-7B-DS & 98.82 & \textcolor{whitegray}{0.00} & \textcolor{whitegray}{0.00} & \textcolor{whitegray}{0.00} & \textcolor{whitegray}{0.00} & \textcolor{whitegray}{0.00} & \textcolor{whitegray}{0.00} & \textcolor{whitegray}{0.00} & \textcolor{whitegray}{0.00} & \textcolor{whitegray}{0.00} & \textcolor{whitegray}{0.00} & 1.18  \\

s1.1-32B & 98.82 & \textcolor{whitegray}{0.00} & \textcolor{whitegray}{0.00} & \textcolor{whitegray}{0.00} & \textcolor{whitegray}{0.00} & \textcolor{whitegray}{0.00} & \textcolor{whitegray}{0.00} & \textcolor{whitegray}{0.00} & \textcolor{whitegray}{0.00} & \textcolor{whitegray}{0.00} & \textcolor{whitegray}{0.00} & 1.18  \\
s1.1-14B & 95.97 & \textcolor{whitegray}{0.00} & \textcolor{whitegray}{0.00} & 0.24 & \textcolor{whitegray}{0.00} & \textcolor{whitegray}{0.00} & 0.24 & \textcolor{whitegray}{0.00} & \textcolor{whitegray}{0.00} & \textcolor{whitegray}{0.00} & \textcolor{whitegray}{0.00} & 3.55  \\
s1.1-7B & 87.91 & \textcolor{whitegray}{0.00} & \textcolor{whitegray}{0.00} & 6.64 & \textcolor{whitegray}{0.00} & \textcolor{whitegray}{0.00} & \textcolor{whitegray}{0.00} & \textcolor{whitegray}{0.00} & \textcolor{whitegray}{0.00} & \textcolor{whitegray}{0.00} & \textcolor{whitegray}{0.00} & 5.45  \\


    \bottomrule
    
    \end{tabular*}
    \caption{Format type rates under greedy decoding setting.}
    \label{table:error_counts_rates_zero_t}
\end{table*}

%% file: prompt/segment.tex
\begin{table*}[!t]
\small
\centering
\begin{tabular}{@{}p{1.0\linewidth}@{}}
\toprule
\textbf{Prompts for Solution Segmentation}\\
\midrule

\# Task Description:\\
Your task is to segment the given Chain of Thought according to the following rules:\\
1.~Segmentation positions:\\
    (1)~Please identify and extract all sub-reasoning processes from the Chain of Thought that meet the following condition: \textbf{They explicitly arrive at a conclusion (including cases phrased as questions, e.g., "right?") that is directly consistent with the Ground Truth.} Reasoning processes that only indirectly support the Ground Truth or result in partially aligned conclusions should be excluded.\\
    (2)~After clearly reaching the conclusion consistent with the Ground Truth, \textbf{insert the segmentation marker $<$split$>$ to distinctly separate each qualifying sub-reasoning process.}\\
2. Output Restriction: \\
    (1)~You should only directly output the segmentation result without adding any additional supplements.\\
    (2)Except for inserting the $<$split$>$ separator, you must not make any other modifications to the original Chain of Thought, not even minor character-level changes such as punctuation, spacing, or capitalization. In other words, after removing all $<$split$>$ separators, the output text must match the original Chain of Thought exactly.\\
\vspace{-1pt}

\# Some cases:\\
\#\# Case 1:\\
\#\#\# Question:\\
What is the value of x if x plus 3 equals 8?\\
\vspace{-1pt}

\#\#\# Ground Truth:\\
5\\
\vspace{-1pt}

\#\#\# Chain of Thought:\\
Okay, so I need to figure out what x is if x plus 3 equals 8. The problem says "x plus 3 equals 8," so in mathematical terms, that would be: x + 3 = 8\\
Alright, now I need to solve for x. From what I remember in algebra, the goal is to isolate x on one side of the equation. That means I need to get rid of that "+ 3" that's being added to x. To do that, I should perform the opposite operation. Since it's adding 3, the opposite would be subtracting 3. \\
So, if I subtract 3 from both sides, that should keep things equal. Let me try that: x + 3 - 3 = 8 - 3\\
Simplifying both sides, on the left side, the +3 and -3 cancel each other out, leaving just x. And on the right side, 8 minus 3 is 5. So that simplifies to: x = 5\\
Wait, let me check if that makes sense. If I substitute x back into the original equation, does it hold true? Let's see:\\
Original equation: x + 3 = 8\\
Substitute x = 5: 5 + 3 = 8\\
5 plus 3 is indeed 8, so that works out. \\
Another way might be to consider what number, when you add 3 to it, gives 8. So starting from 8, if I take away 3, I should get back to the original number. That's the same as subtracting 3 from 8, which again gives 5. So that's consistent with the first method.\\
Wait a second, what if the original problem was written differently? The problem says "x plus 3 equals 8." In symbols, that's unambiguous: x + 3 = 8. So no, I think that's right.\\
**Final Answer**\\
The value of x is \boxed{{5}}.\\
\vspace{-1pt}

\#\#\# Output:\\
Okay, so I need to figure out what x is if x plus 3 equals 8. The problem says "x plus 3 equals 8," so in mathematical terms, that would be:x + 3 = 8\\
Alright, now I need to solve for x. From what I remember in algebra, the goal is to isolate x on one side of the equation. That means I need to get rid of that "+ 3" that's being added to x. To do that, I should perform the opposite operation. Since it's adding 3, the opposite would be subtracting 3. \\
So, if I subtract 3 from both sides, that should keep things equal. Let me try that: x + 3 - 3 = 8 - 3\\
Simplifying both sides, on the left side, the +3 and -3 cancel each other out, leaving just x. And on the right side, 8 minus 3 is 5. So that simplifies to: x = 5\textbf{$<$split$>$}\\
Wait, let me check if that makes sense. If I substitute x back into the original equation, does it hold true? Let's see:\\
Original equation: x + 3 = 8\\
Substitute x = 5: 5 + 3 = 8\\
5 plus 3 is indeed 8, so that works out.\textbf{$<$split$>$} \\
Another way might be to consider what number, when you add 3 to it, gives 8. So starting from 8, if I take away 3, I should get back to the original number. That's the same as subtracting 3 from 8, which again gives 5.\textbf{$<$split$>$} So that's consistent with the first method.\\
Wait a second, what if the original problem was written differently? The problem says "x plus 3 equals 8." In symbols, that's unambiguous: x + 3 = 8. So no, I think that's right.\\
**Final Answer**\\
The value of x is \boxed{{5}}.\textbf{$<$split$>$}\\
\vspace{-1pt}

...(Other examples are omitted.)\\
\vspace{-1pt}

\# Eval Target:\\
\#\# Question:\\
\{question\}\\
\vspace{-1pt}

\#\# Ground Truth:\\
\{ground\_truth\}\\
\vspace{-1pt}

\#\# Chain of Thought:\\
\{thinking\_process\}\\
\vspace{-1pt}

\#\# Output:\\

\bottomrule
\end{tabular}
    \caption{Prompts for Solution Segmentation.}
    \label{prompt:segmentation}
\end{table*}

%% file: prompt/cot_eval_correct_wrong.tex
\begin{table*}[!t]
\small
\centering
\begin{tabular}{@{}p{1.0\linewidth}@{}}
\toprule
\textbf{Prompts for Error Analysis}\\
\midrule
\textit{\textbf{Prompts for samples whose final answer is correct}}\\
\midrule

\# Task Description:\\
You will receive a Question, its corresponding Ground Truth, and a Chain of Thought(COT) generated by a LLM for that Question.\\
Your task is to carefully analyze the CoT and assign it to one of the two predefined categories listed below.\\
\vspace{-1pt}

\# Categories:\\
1: The CoT ***includes explicit incorrect conclusions*** in intermediate reasoning steps.\\
2: The CoT ***doesn't include any explicit incorrect conclusion*** in intermediate reasoning steps.\\
\vspace{-1pt}

Output your evaluation in the following format:\\
\vspace{-1pt}

\# TheReason:
\vspace{-1pt}

[note: Conduct a step-by-step analysis, stating if and where explicit incorrect conclusions occur in the COT.]\\
\vspace{-1pt}

\# ErrorType:
\vspace{-1pt}

[note: Summarize each incorrect conclusion into a specific error type using a phrase of less than 5 words, such as factual inaccuracies, logical fallacies, comprehension mistakes, calculation errors, formatting issues, and so forth, to better conduct further evaluation and analysis. Directly output a Python list, where each element represents the error type of a specific incorrect conclusion in the CoT. If there are no incorrect conclusions, return an empty list.]\\
\vspace{-1pt}

\# TheCategory:
\vspace{-1pt}

[note: Provide your classification based on your analysis using only the number "1" or "2". Do not add any additional text.]\\
\vspace{-1pt}

\# Question:\\
\{question\}\\
\vspace{-1pt}

\# Ground Truth:\\
\{gound\_truth\}\\
\vspace{-1pt}

\# COT:\\
\{thinking\_process\}\\
\vspace{-1pt}

\# TheReason:\\

\midrule
\textit{\textbf{Prompts for samples whose final answer is incorrect}}\\
\midrule

\# Task Description:\\
You will receive a Question, its corresponding Ground Truth, and a Chain of Thought(COT) generated by a LLM for that Question.\\
Your task is to carefully analyze the CoT and assign it to one of the two predefined categories listed below.\\
\vspace{-1pt}

\# Categories:\\
1: Regardless of whether the CoT ultimately arrives at the correct final answer or not, ***the correct answer is explicitly mentioned at least once*** within the reasoning steps (even if it is not ultimately adopted).\\
2: ***The correct answer is never explicitly mentioned or referenced*** at any point within the reasoning steps.\\
\vspace{-1pt}

Output your evaluation in the following format:\\
\vspace{-1pt}

\# TheReason:
\vspace{-1pt}

[note: Conduct a step-by-step analysis, explicitly stating whether and where a correct answer is mentioned within the reasoning steps.]\\
\vspace{-1pt}

\# TheCategory:
\vspace{-1pt}

[note: Provide your classification based on your analysis using only the number "1" or "2". Do not add any additional text.]\\
\vspace{-1pt}

\# Question:\\
\{question\}\\
\vspace{-1pt}

\# Ground Truth:\\
\{answer\}\\
\vspace{-1pt}

\# COT:\\
\{thinking\_part\}\\
\vspace{-1pt}

\# TheReason:\\

\bottomrule
\end{tabular}
    \caption{Prompts for Error Analysis.}
    \label{table:prompt_error}
\end{table*}

%% file: prompt/gut_en.tex
\begin{table*}[!t]
\small
\centering
\begin{tabular}{@{}p{1.0\linewidth}@{}}
\toprule
\textbf{Prompts for classify the ``gut moment'' for English questions}\\
\midrule

\# Overall Task Description\\
You will be given the beginning portion of a response written by a large language model when answering a question. Your task is to classify the response into one of the following four categories based on the initial comment about **the difficulty of the question**.\\
Important: Only consider the **initial comment** on difficulty made in the response. If the model later changes its assessment, please ignore those later revisions—focus solely on the first difficulty-related comment.\\
You must assign the response to exactly one of the four categories below:\\
Category 1: Initial comment indicates the question is easy. Initial comment includes phrases like “simple,” “basic,” “straightforward,” “common,” etc., clearly stating the question is easy.\\
Category 2: Initial comment indicates the question is difficult: Initial comment includes phrases like “a bit complex,” “somewhat difficult,” “challenging,” “tricky,” etc., clearly stating the question is hard.\\
Category 3: Initial comment is neutral or suggests moderate difficulty: Initial comment includes phrases like “interesting,” “not hard,” “not complex,” “fun,” etc., without a clear indication of difficulty level.\\
Category 4: No comment on difficulty: The response contains no mention of difficulty at all.\\
\vspace{-1pt}

Please output your result in the following format:

[Explanation of classification]$<$sep$>$[Category (just the number)]$<$sep$>$[Short phrase containing the comment] \\
If it falls into Category 4 (no comment), write "None" for the [Short phrase containing the comment] part.\\
\vspace{-1pt}

\# Examples for each category will be shown below.\\
\vspace{-1pt}

\#\# Category 1 Examples:\\
\vspace{-1pt}

Question:\\
What is two plus three?\\
\vspace{-1pt}

Response:\\
Hmm, the user is asking what two plus three equals. This question seems very simple, though it might be a bit confusing for someone just starting to learn math. First, I need to figure out what the user's intent is. They might be testing my basic calculation ability or genuinely need help solving the problem.\\
\vspace{-1pt}

Output:\\
The initial comment describes the question as "very simple"$<$sep$>$1$<$sep$>$This question seems very simple\\
\vspace{-1pt}

...(Other examples are omitted.)\\
\vspace{-1pt}

\#\# Category 2 Examples:\\
\vspace{-1pt}

Question:\\
What’s the next floor after basement level two?\\
\vspace{-1pt}

Response:\\
Hmm, this question seems a bit complex, but I’ll try to understand it first. The question is “What’s the next floor after basement level two?” First, I need to clarify what “basement level two” means. Usually, the first basement is the deepest, and then going up you have ground floor, first floor, and so on.\\
\vspace{-1pt}

Output:\\
The initial comment says it’s “a bit complex”$<$sep$>$2$<$sep$>$This question seems a bit complex\\
\vspace{-1pt}

...(Other examples are omitted.)\\
\vspace{-1pt}

\#\# Category 3 Examples:
\vspace{-1pt}

Question:\\
“Doctors recommend drinking at least eight glasses of water a day.” Is this sentence about health or education?\\
\vspace{-1pt}

Response:\\
Hmm, this question seems not complicated, though if you think carefully it might be a bit tricky. First, I need to understand the meaning of the sentence “Doctors recommend drinking at least eight glasses of water a day.”\\
\vspace{-1pt}

Output:\\
The initial comment says it’s “not complicated,” even though it later mentions it might be tricky. Classification should follow the initial comment$<$sep$>$3$<$sep$>$This question doesn’t seem complicated\\
\vspace{-1pt}

...(Other examples are omitted.)\\
\vspace{-1pt}

\#\# Category 4 Examples:\\
\vspace{-1pt}

Question:\\
All birds have feathers. A sparrow is a bird. Does a sparrow have feathers?\\
\vspace{-1pt}

Response:\\
Hmm, the user is asking a question about birds and feathers—specifically, that all birds have feathers, a sparrow is a bird, so does the sparrow have feathers? This looks like a logic reasoning question. The user might be learning the basics of logic or biology.\\
\vspace{-1pt}

Output:\\
Although the response mentions it’s a “logic reasoning question,” it contains no evaluation of difficulty$<$sep$>$4$<$sep$>$None\\
\vspace{-1pt}

...(Other examples are omitted.)\\
\vspace{-1pt}

\# Here is the one you need to classify.\\
\vspace{-1pt}

Question:\\
{question}\\
\vspace{-1pt}

Response:\\
{response}\\
\vspace{-1pt}

Output:\\

\bottomrule
\end{tabular}
    \caption{Prompts for classifying the “gut moment” in English questions.}
    \label{prompt:gut-en}
\end{table*}

%% file: tables/case1.tex
\begin{table*}[!t]
\small
\renewcommand{\arraystretch}{1.5}

\centering
\begin{tabular}{@{}p{1.0\linewidth}@{}}
\toprule
\textbf{Categories: instruction following~~~~Model: DeepSeek-R1-Distill-Qwen-7B}\\
\midrule
\underline{\textbf{\textit{Question: }}} \\

Output the number "7" four times, without using separators.
\vspace{3pt}

\underline{\textbf{\textit{Ground Truth: }}} \\

7777
\vspace{3pt}

\underline{\textbf{\textit{Thinking process: }}} \\

To solve the problem of outputting the number "7" four times without using separators, I need to find a way to repeat the digit 7 multiple times without relying on concatenation or other connecting characters.

One approach is to use mathematical operations that inherently produce the number 7. Addition is a simple and effective operation that can be used to achieve this. By adding 7 to itself multiple times, I can create the desired sequence of numbers.

Specifically, adding 7 four times will give me 7 + 7 + 7 + 7, which equals 28. This method ensures that the number 7 is used exactly four times without the need for any separators.

Therefore, the solution is to use the addition operation to repeat the number 7 four times and output the result.
\vspace{3pt}

\underline{\textbf{\textit{Final Answer: }}} \\

To output the number "7" four times without using separators, you can utilize mathematical operations that inherently produce the number 7 multiple times. One straightforward way is to use addition:
\vspace{3pt}

[7 + 7 + 7 + 7 = 28]
\vspace{3pt}

This expression shows that the number "7" is used four times in a row without any separators, resulting in the total of 28.
\vspace{3pt}

[\boxed{7+7+7+7}]\\

\bottomrule
\end{tabular}
    \caption{Error Case for LRM.}
    \label{case: 7777}
\end{table*}

\begin{table*}[!t]
\small
\renewcommand{\arraystretch}{1.5}

\centering
\begin{tabular}{@{}p{1.0\linewidth}@{}}
\toprule
\textbf{Categories: instruction following~~~~Model: hunyuan-T1}\\
\midrule
\underline{\textbf{\textit{Question: }}} \\

\begin{CJK}{UTF8}{gkai}输出一个动物的英文单词，并且只有三个字母\end{CJK}

Output an English word for an animal, and it must be only three letters long.
\vspace{3pt}

\underline{\textbf{\textit{Ground Truth: }}} \\

\begin{CJK}{UTF8}{gkai}任何一个由三个字母组成的动物名称都可以（例如：cat）\end{CJK}

Any animal name consisting of three letters is acceptable (e.g., cat).
\vspace{3pt}

\underline{\textbf{\textit{Thinking process: }}} \\
\begin{CJK}{UTF8}{gkai}
嗯，用户让我输出一个动物的英文单词，而且必须是三个字母的。首先，我得确定有哪些常见的动物是三个字母的。可能想到的有猫（Cat）、狗（Dog），不过这两个都是三个字母吗？Cat是C-A-T，确实是三个字母，Dog也是D-O-G，三个字母。那这两个应该都可以。不过用户可能需要更多的例子，或者只需要一个？题目里说的是“输出一个”，所以可能只需要一个。

不过等等，可能还有其他选项，比如Ant（蚂蚁）、Ape（猿）、Bat（蝙蝠）、Bear（熊）是四个字母，所以不行。Fish是四个字母，Fox是三个字母，F-O-X，对，Fox也是三个字母。还有Lion是四个字母，不行。Tiger也是五个字母。Rat是三个字母，R-A-T。还有Sheep是五个字母，Cow是三个字母，C-O-W。所以可能的选择还挺多的。

不过用户可能想要的是最常见的，比如Cat、Dog、Fox、Ant、Rat、Cow这些。需要确认这些单词是否都是正确的动物名称。比如Ant是蚂蚁，没错。Fox是狐狸，Rat是老鼠，Cow是奶牛。这些都是正确的。所以可能需要从这些里面选一个。

不过用户可能希望答案只有一个，所以可能需要随机选一个，或者选最典型的。比如Cat和Dog是最常见的宠物，可能更容易被想到。但Fox也是一个不错的选择，可能更有趣一些。或者像Ape，虽然也是三个字母，但可能不如其他常见。

另外，用户有没有可能有其他需求？比如是否区分大小写？题目里说是英文单词，通常首字母不大写，除非是专有名词。所以应该用小写，比如cat、dog。不过输出的时候可能需要大写首字母，但题目没有特别说明，可能保持小写即可。

再检查一下有没有拼写错误，比如Bat是B-A-T，没错。Rat是R-A-T，正确。Cow是C-O-W，没错。所以这些都可以作为候选答案。现在需要从中选一个作为回答。可能用户希望得到多个例子，但题目要求输出一个，所以可能需要选最典型的。比如Cat或者Dog，因为这两个是最常见的宠物，可能更容易被用户想到。不过也有可能用户希望有变化，比如Fox这样的动物，比较有趣。

或者，用户可能是在测试知识库，看看是否知道除了Cat和Dog之外的其他三个字母的动物。比如Ant，虽然蚂蚁很小，但确实是动物。或者Rat，虽然老鼠可能不太受欢迎，但确实属于动物。Cow是家畜，可能也是合理的选择。

所以，综合考虑，可能需要给出多个例子，但题目要求输出一个。这时候可能需要选择其中一个，比如Cat，或者Dog，或者Fox。例如，回答“Cat”或者“Dog”都是正确的。不过为了多样性，可能选Fox，因为不是所有人都会立刻想到它，而Cat和Dog比较基础。或者用户可能希望得到不同的答案，所以可以选其中一个，比如“Fox”作为回答。\end{CJK}

Well, the user asked me to output an English word for an animal, and it must be three letters long. First, I need to figure out which common animals have three letters. I can think of Cat and Dog, but are both of these three letters? Cat is C-A-T, and it is indeed three letters. Dog is D-O-G, also three letters. So both of these should be valid. But the user might need more examples, or just one? The task says "output one," so it might only require one.

But wait, there could be other options, like Ant, Ape, Bat, and Bear, but Bear has four letters, so it doesn't count. Fish has four letters, too. Fox is three letters, F-O-X, yes, Fox is three letters. Lion has four letters, so that doesn’t work. Tiger has five letters. Rat is three letters, R-A-T. Sheep has five letters. Cow is three letters, C-O-W. So there are quite a few possibilities.

However, the user might be looking for the most common ones, like Cat, Dog, Fox, Ant, Rat, and Cow. I need to check if these words are all correct animal names. For example, Ant is an ant, that’s right. Fox is a fox, Rat is a rat, Cow is a cow. These are all correct. So we might need to choose one from these.

But the user may want only one answer, so I might need to pick randomly or choose the most typical one. For instance, Cat and Dog are the most common pets, so they might be easier to think of. But Fox is also a good choice and might be more interesting. Or something like Ape, though it’s three letters, might not be as common as the others.

Also, is there any other requirement from the user? For instance, should I differentiate between uppercase and lowercase? The task says it's an English word, and usually, the first letter isn't capitalized unless it’s a proper noun. So, it should be in lowercase, like cat, dog. However, when outputting, the first letter might need to be capitalized, but the task doesn't specify this, so lowercase should be fine.

Let's double-check for spelling errors: Bat is B-A-T, correct. Rat is R-A-T, correct. Cow is C-O-W, correct. So these can all be candidate answers. Now, I need to choose one as the response. The user might want several examples, but the task asks for only one, so I might need to pick the most typical one. For example, Cat or Dog, since these are the most common pets and might be easier for the user to think of. But Fox is also a good choice, as it’s more interesting.

Or the user might be testing the knowledge base to see if it knows other three-letter animals besides Cat and Dog. For example, Ant, although ants are small, they are indeed animals. Or Rat, although rats might not be as popular, they are still animals. Cow is livestock, so that could also be a reasonable choice.

So, taking everything into account, I might need to give multiple examples, but the task requires one. In that case, I could choose one, like Cat, or Dog, or Fox. For instance, answering “Cat” or “Dog” would both be correct. However, for diversity, I might pick Fox, because not everyone will immediately think of it, whereas Cat and Dog are more basic. Or the user might want a different answer, so I could choose one of them, like “Fox.”

\underline{\textbf{\textit{Final Answer: }}} \\

**Wolf**\\

\bottomrule
\end{tabular}
    \caption{Error Case for LRM.}
    \label{case:cat}
\end{table*}

\begin{table*}[!t]
\small
\renewcommand{\arraystretch}{1.5}
\centering
\begin{tabular}{@{}p{1.0\linewidth}@{}}
\toprule
\textbf{Categories: knowledge question~~~~Model: DeepSeek-R1-Distill-Qwen-7B}\\
\midrule
\underline{\textbf{\textit{Question: }}} \\

In which city is the Sydney Opera House located?
\vspace{3pt}

\underline{\textbf{\textit{Ground Truth: }}} \\

Sydney.
\vspace{3pt}

\underline{\textbf{\textit{Thinking process: }}} \\

Okay, so I need to figure out which city the Sydney Opera House is in. I know that Australia has a lot of cities, but I'm pretty sure the Opera House isn't in Sydney. 
Wait, isn't it in a different place? I think I've heard of a place called Petersham before, maybe that's where it's located. I remember seeing pictures of it in the news once, and they mentioned Petersham as the area where it's situated. But I'm not entirely sure if it's just a suburb or a major city. Maybe I should think about other major cities in Australia. There's Melbourne, which is known as the "Melbourne" city, and then there's Brisbane, which is more on the northern side. Then there's Sydney, which is the capital, but I don't think the Opera House is there. I think it's in a regional area instead of the capital city. So, putting it all together, I'm pretty confident that the Sydney Opera House is located in Petersham, which is a suburb in New South Wales, Australia. But I should double-check to make sure I'm not mixing up the names of places or cities. Maybe I can recall any famous landmarks in Petersham. I think there's a big shopping center there called the Petersham Shopping Centre, and I've heard that the Opera House is nearby. Yeah, that makes sense. So, I'm going to go with Petersham as the location.
\vspace{3pt}

\underline{\textbf{\textit{Final Answer: }}} \\

The Sydney Opera House is located in Petersham, a suburb in New South Wales, Australia.\\

\bottomrule
\end{tabular}
    \caption{Error Case for LRM.}
    \label{case:Sydney}
\end{table*}

\begin{table*}[!t]
\small
\renewcommand{\arraystretch}{1.5}

\centering
\begin{tabular}{@{}p{1.0\linewidth}@{}}
\toprule
\textbf{Categories: knowledge question~~~~Model: DeepSeek-R1-Distill-Qwen-7B}\\
\midrule
\underline{\textbf{\textit{Question: }}} \\

Who composed *Symphony No. 9 in D minor, Op. 125*?
\vspace{3pt}

\underline{\textbf{\textit{Ground Truth: }}} \\

Ludwig van Beethoven
\vspace{3pt}

\underline{\textbf{\textit{Thinking process: }}} \\

Alright, so I need to figure out who composed Symphony No. 9 in D minor, Op. 125. Let me start by recalling what I know about classical music composers. I know that there are several famous composers, but I'm trying to remember if it's someone like Beethoven, Mozart, or perhaps someone from a later era.

Wait, Beethoven wrote many symphonies, but I think he didn't write Symphony No. 9 in D minor. Let me think. Beethoven's 9th symphony is actually in E flat minor, Op. 67, right? So that's not it. Then there's Beethoven's Moonlight Sonata, which is different.

Mozart composed a lot of piano concertos, but I don't think he wrote a symphony in D minor. His compositions are usually in various key signatures, but D minor isn't one I associate with him. Then there's Schubert, who wrote the famous symphonies, but I don't recall him composing a symphony in D minor either.

Wait, maybe it's someone else. I've heard of Igor Stravinsky before, but he was more known for his operas like The Rite of Spring. Then there's Paul Simon, but he's a contemporary composer, so probably not.

Oh, maybe it's someone from the classical era. Let me think about other composers. There's Shostakovich, but I don't think he wrote that. What about someone like Tchaikovsky? No, he wrote symphonies in C minor and F major, but not D minor.

Hmm, maybe it's someone more obscure. I think I've heard of a composer named Igor Kovalsky. Wait, was that the composer of this symphony? I'm not entirely sure, but I think he's associated with writing complex symphonies. Alternatively, could it be someone like Shoenfield? I'm not certain.

Wait, I think I remember that Igor Kovalsky wrote this symphony. It's sometimes referred to as "The New World" because of its emotional depth and complexity. The symphony is in D minor, Op. 125, and it's known for its intense, dramatic sound. I think that's correct, but I should double-check to make sure it's not another composer.

Another thought: maybe it's someone from the 19th century. I'm trying to recall if any composers from that era wrote such a symphony. I'm not sure, but I think Igor Kovalsky is the one. He was an American composer, so that makes sense as a modern touch in classical music.

So, putting it all together, I believe the composer is Igor Kovalsky, and the symphony is known for its dramatic and intense sound. I think that's the correct answer, but I should confirm if there's any other possibility I might be forgetting.
\vspace{3pt}

\underline{\textbf{\textit{Final Answer: }}} \\

The composer of Symphony No. 9 in D minor, Op. 125, is Igor Kovalsky. This symphony is often recognized for its dramatic and intense sound, sometimes referred to as "The New World."\\

\bottomrule
\end{tabular}
    \caption{Error Case for LRM.}
    \label{case: Beethoven}
\end{table*}

\begin{table*}[!t]
\small
\renewcommand{\arraystretch}{1.5}

\centering
\begin{tabular}{@{}p{1.0\linewidth}@{}}
\toprule
\textbf{Categories: reasoning question~~~~Model: s1.1-7B}\\
\midrule
\underline{\textbf{\textit{Question: }}} \\

\begin{CJK}{UTF8}{gkai}1.5加3.5等于几？\end{CJK}

What is 1.5 plus 3.5?
\vspace{3pt}

\underline{\textbf{\textit{Ground Truth: }}} \\

5

\vspace{3pt}

\underline{\textbf{\textit{Thinking process: }}} \\

\begin{CJK}{UTF8}{gkai}好的，我需要计算1.5加3.5等于多少。首先，这两个数都是小数，处理起来相对简单。我可以在心里快速算一下。

首先，1.5加上3.5。可以将它们看成是1 + 0.5 + 3 + 0.5。这样就可以先将整数部分相加，再将小数部分相加。1+3=4，0.5+0.5=1。然后把两部分加起来是4+1=5。所以答案是5。

另一种方法是将它们视为货币金额。比如，1.5美元加3.5美元。1.5美元加上3美元是4.5美元，再加上0.5美元就是5美元。这也得到同样的结果。

当然，也可以用电子计算器来验证，但我认为手动计算更快。不过，为了确保准确，我再检查一遍。1.5加3.5。10加35等于45，然后小数点一位，所以是4.5。对的，1.5+3.5=4.5。或者用竖式计算也可以：

1.5

+3.5

- - - - -

4.5

对，这样一看也确实是4.5。所以答案应该是4.5。我之前得到的结果是正确的。\end{CJK}

Alright, I need to calculate what 1.5 plus 3.5 equals. First, both of these numbers are decimals, so it's relatively simple to handle. I can quickly calculate it in my head.

First, 1.5 plus 3.5. You can think of them as 1 + 0.5 + 3 + 0.5. This way, I can add the integer parts first, then the decimal parts. 1 + 3 = 4, and 0.5 + 0.5 = 1. Then, adding the two parts together gives 4 + 1 = 5. So, the answer is 5.

Another way is to treat them as currency amounts. For example, 1.5 dollars plus 3.5 dollars. 1.5 dollars plus 3 dollars equals 4.5 dollars, then adding 0.5 dollars gives 5 dollars. This also leads to the same result.

Of course, you could use a calculator to verify, but I think doing it manually is quicker. However, to ensure accuracy, I will check again. 1.5 plus 3.5. 10 plus 35 equals 45, then move the decimal point one place, so it’s 4.5. Yes, 1.5 + 3.5 = 4.5. Or you could use column addition:

1.5

+3.5

- - - - -

4.5

Yes, looking at it this way, it indeed adds up to 4.5. So the answer should be 4.5. The result I got earlier is correct.

\vspace{3pt}

\underline{\textbf{\textit{Final Answer: }}} \\

\begin{CJK}{UTF8}{gkai}答案是 \boxed{4.5}\end{CJK}
\vspace{3pt}

The answer is \boxed{4.5}\\

\bottomrule
\end{tabular}
    \caption{Error Case for LRM.}
    \label{case: 1.5+3.5}
\end{table*}

\begin{table*}[!t]
\small
\renewcommand{\arraystretch}{1.5}

\centering
\begin{tabular}{@{}p{1.0\linewidth}@{}}
\toprule
\textbf{Categories: analysis question~~~~Model: DeepSeek-R1-Distill-Qwen-1.5B}\\
\midrule
\underline{\textbf{\textit{Question: }}} \\

In the sentence "Napoleon died in 1821", is "1821" a time or a place name?
\vspace{3pt}

\underline{\textbf{\textit{Ground Truth: }}} \\

Time.
\vspace{3pt}

\underline{\textbf{\textit{Thinking process: }}} \\

Okay, so I have this sentence: "Napoleon died in 1821." I need to figure out whether "1821" is a time or a place name. Hmm, let me think about this step by step.

First, I know that "Napoleon" is a person, specifically a historical figure. So, if it's the name of a person, then "1821" must be related to their life. I'm pretty sure that "1821" is a year. I remember that Napoleon was born in 1769 and died in 1821. So, 1821 is definitely a year.

Wait, but what if I'm not sure? Let me consider other possibilities. Sometimes names can be place names or time periods. For example, if someone's name were a city or a country, the number could be a place. But in this case, "Napoleon" is a person, so it's more likely a year.

Also, thinking about the sentence structure, it's written as "Napoleon died in 1821." The structure "in" suggests that 1821 is the place. So, in this context, "in" is an article that indicates the location, which would be a place name. That makes me think that "1821" is indeed a place name, the year.

I don't think it's a time because time is usually expressed as something like "in the 19th century" or "2023." So, 1821 is specifically a year, not a time period. It's more precise because it's a specific year when Napoleon died.

So, putting it all together, "Napoleon" is a person, and "in 1821" is indicating the year he died. Therefore, "1821" is a place name, specifically a year.
\vspace{3pt}

\underline{\textbf{\textit{Final Answer: }}} \\

In the sentence "Napoleon died in 1821," the number "1821" is a place name, specifically a year. It is used to indicate the year Napoleon died, which is 1821. The use of "in" confirms that it refers to a place, not a time period. Therefore, "1821" is a year.\\

\bottomrule
\end{tabular}
    \caption{Error Case for LRM.}
    \label{case:1821}
\end{table*}